\documentclass{article}

\usepackage{arxiv}

\usepackage[T1]{fontenc}
\usepackage[utf8]{inputenc}
\usepackage{authblk}
\usepackage{hyperref}       
\usepackage{url}            
\usepackage{booktabs}       
\usepackage{amsfonts}       
\usepackage{nicefrac}       
\usepackage{microtype}      
\usepackage{lipsum}
\usepackage{fancyhdr}       
\usepackage{graphicx}       
\graphicspath{{media/}}     

\usepackage{graphicx}%
\usepackage{subfigure}
\usepackage{multirow}%
\usepackage{amsmath,amssymb,amsfonts}%
\usepackage{amsthm}%
\usepackage{mathrsfs}%
\usepackage[title]{appendix}%
\usepackage{xcolor}%
\usepackage{textcomp}%
\usepackage{manyfoot}%
\usepackage{booktabs}%
\usepackage{algorithm}%
\usepackage{algorithmicx}%
\usepackage{algpseudocode}%
\usepackage{listings}%
\usepackage{todonotes}

\usepackage{makecell}

\usepackage{colortbl}
\usepackage{xcolor}
\definecolor{tabgray}{gray}{0.90}

\usepackage{comment}
\usepackage{soul}
\usepackage{mathtools}

\usepackage{ulem}
\usepackage{longtable}
\usepackage{multirow}
\usepackage{array}
\newcolumntype{H}{>{\setbox0=\hbox\bgroup}c<{\egroup}@{}}

\usepackage{bbding}

\newtheorem{proposition}{Proposition}%
\newenvironment{propref}[1]
  {\renewcommand{\theproposition}{\ref{#1}$'$}%
   \addtocounter{proposition}{-1}%
   \begin{proposition}}
  {\end{proposition}}

\usepackage{colortbl}
\usepackage{xcolor}
\definecolor{tabgray}{gray}{0.90}

\title{Revisiting Deep Ensemble for Out-of-Distribution Detection: A Loss Landscape Perspective
}

\author[1]{Kun Fang}
\author[2]{Qinghua Tao}
\author[1]{Xiaolin Huang}
\author[1]{Jie Yang}
\affil[1]{Department of Automation, Shanghai Jiao Tong University\\
{\tt\small
\{fanghenshao,xiaolinhuang,jieyang\}@sjtu.edu.cn}}

\affil[2]{ESAT-STADIUS, KU Leuven, Belgium\\
{\tt\small qinghua.tao@esat.kuleuven.be}}

\begin{document}
\maketitle

\begin{abstract}
Existing Out-of-Distribution (OoD) detection methods address to detect OoD samples from In-Distribution (InD) data mainly by exploring differences in features, logits and gradients in Deep Neural Networks (DNNs).
We in this work propose a new perspective upon \textit{loss landscape} and \textit{mode ensemble} to investigate  OoD detection. 
In the optimization of DNNs, there exist many  local optima in the parameter space, or namely \textit{modes}.
Interestingly, we observe that these independent modes, which all reach low-loss regions with InD data (training and test data), yet yield significantly different loss landscapes with OoD data.
Such an observation provides a novel view to investigate the OoD detection from the loss landscape, and further suggests significantly fluctuating OoD detection performance across these modes.
For instance, FPR values of the RankFeat \cite{song2022rankfeat} method can range from 46.58\% to 84.70\% among 5 modes, showing  uncertain detection performance evaluations across independent modes.
Motivated by such diversities on OoD loss landscape across modes, we revisit the deep ensemble method for OoD detection through mode ensemble, leading to improved performance and benefiting the OoD detector with reduced variances.
Extensive experiments covering varied OoD detectors and network structures illustrate high variances across modes and validate the superiority of mode ensemble in boosting OoD detection.
We hope this work could attract attention in the view of independent modes in the loss landscape of OoD data and more reliable evaluations on OoD detectors.
\end{abstract}


\section{Introduction}
\label{sec:intro}
For years Deep Neural Networks (DNNs) have shown powerful abilities in fitting In-Distribution (InD) data.
Given a DNN $f$ well-trained on data from some distribution ${\cal P}_{\rm in}$,  $f$ also performs well on the test data, which is commonly regarded as being from the same distribution ${\cal P}_{\rm in}$ as the training data \cite{zhang2021understanding}.
However, the generalization ability of $f$ on the data from an out distribution ${\cal P}_{\rm out}$ is still prohibitive, which brings a large number of researches \cite{hendrycks2016baseline,liang2018enhancing,lee2018simple,liu2020energy,lee2020gradients,sun2021react,huang2021importance,song2022rankfeat,zhu2022boosting,sun2022out,wang2022vim,yu2023block} trying to detect Out-of-Distribution (OoD) samples in the inference stage, known as the popular OoD detection problem \cite{yang2024generalized,shen2021towards}.

\begin{figure*}[t]
    \centering
    
    \subfigure[Training set of InD data.]{
    \label{fig:intro-loss-landscape-1}
    \includegraphics[width=0.32\linewidth]{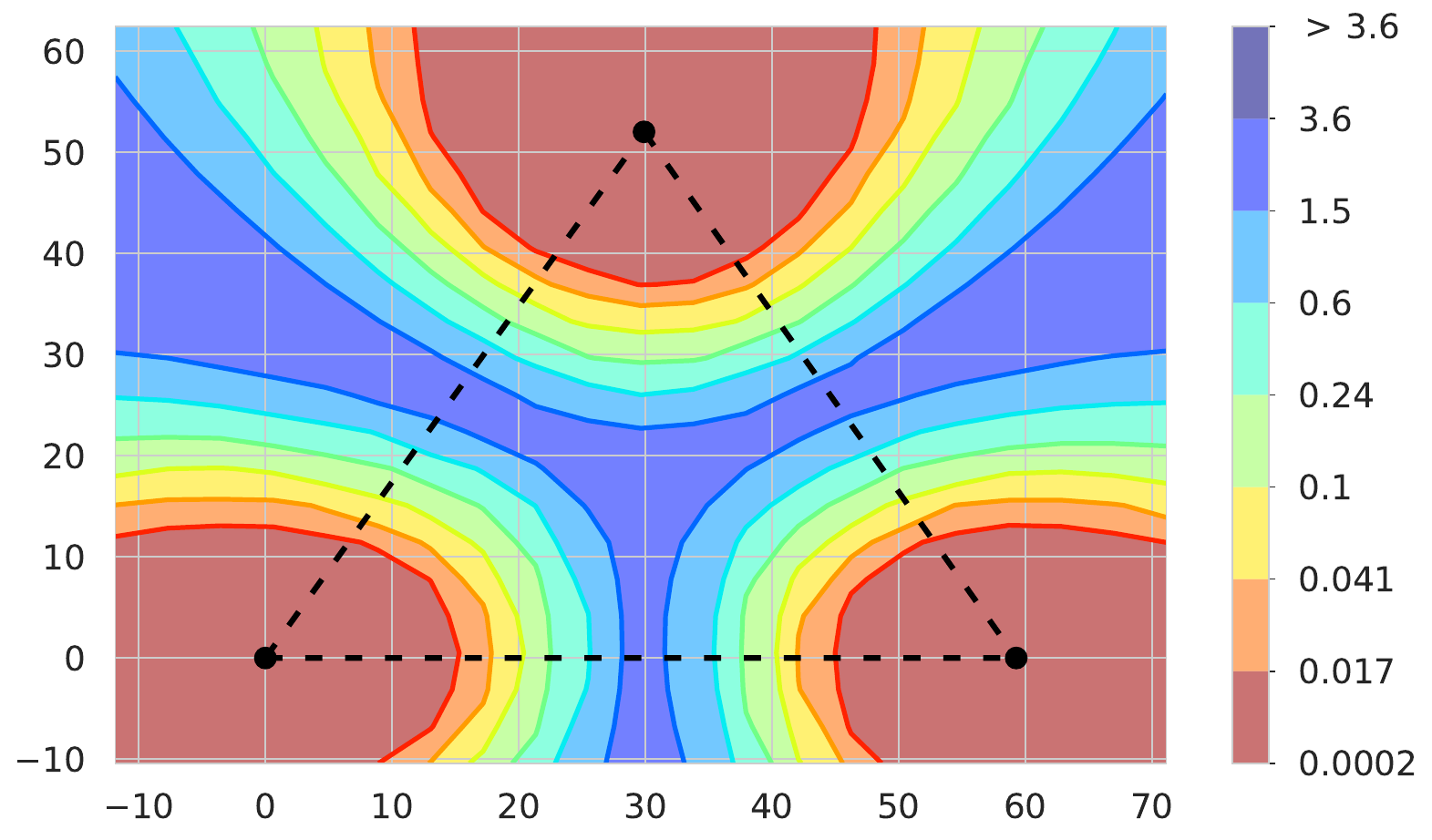}}
    \subfigure[Test set of InD data.]{
    \label{fig:intro-loss-landscape-2}
    \includegraphics[width=0.32\linewidth]{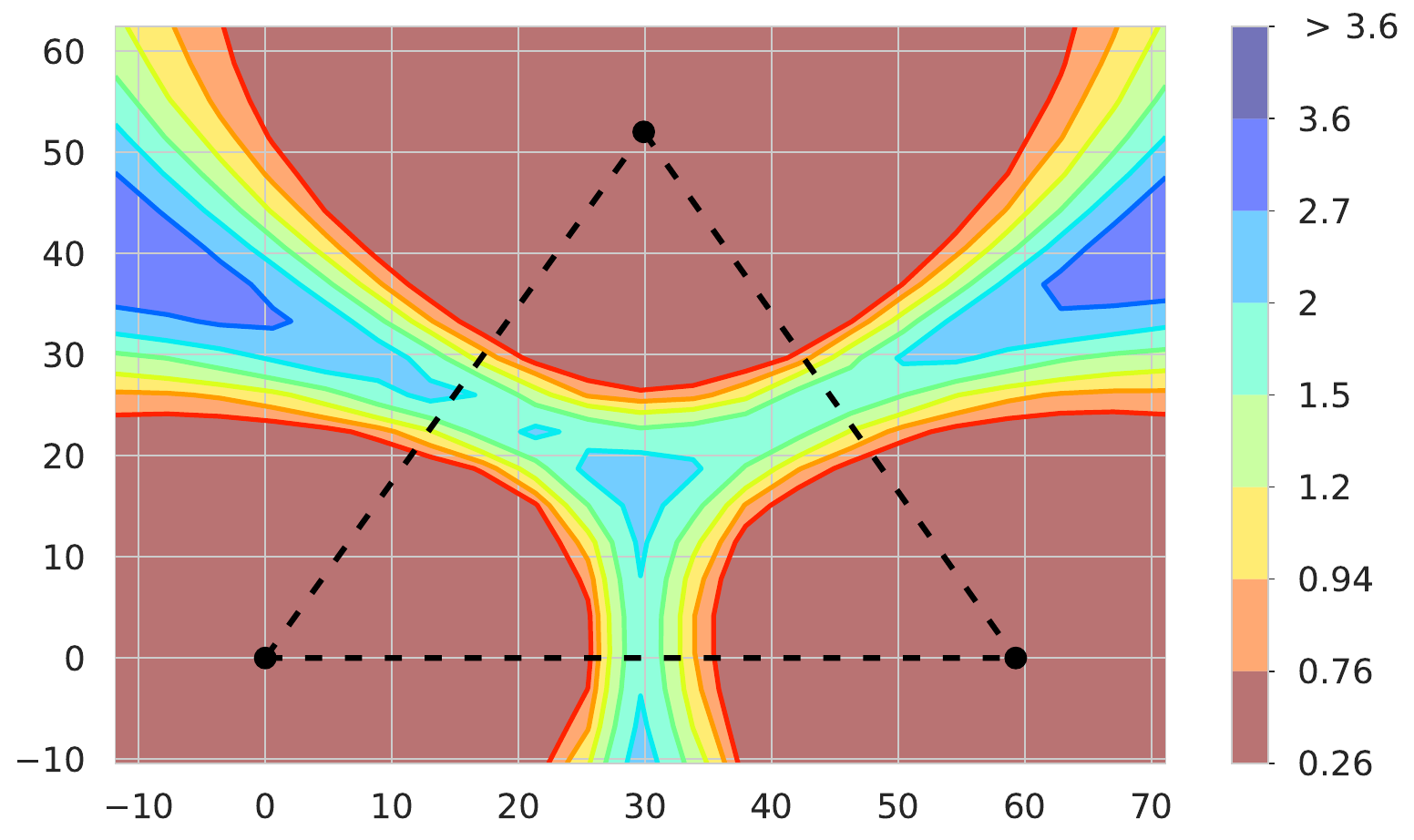}}
    \subfigure[OoD data set.]{
    \label{fig:intro-loss-landscape-3}
    \includegraphics[width=0.32\linewidth]{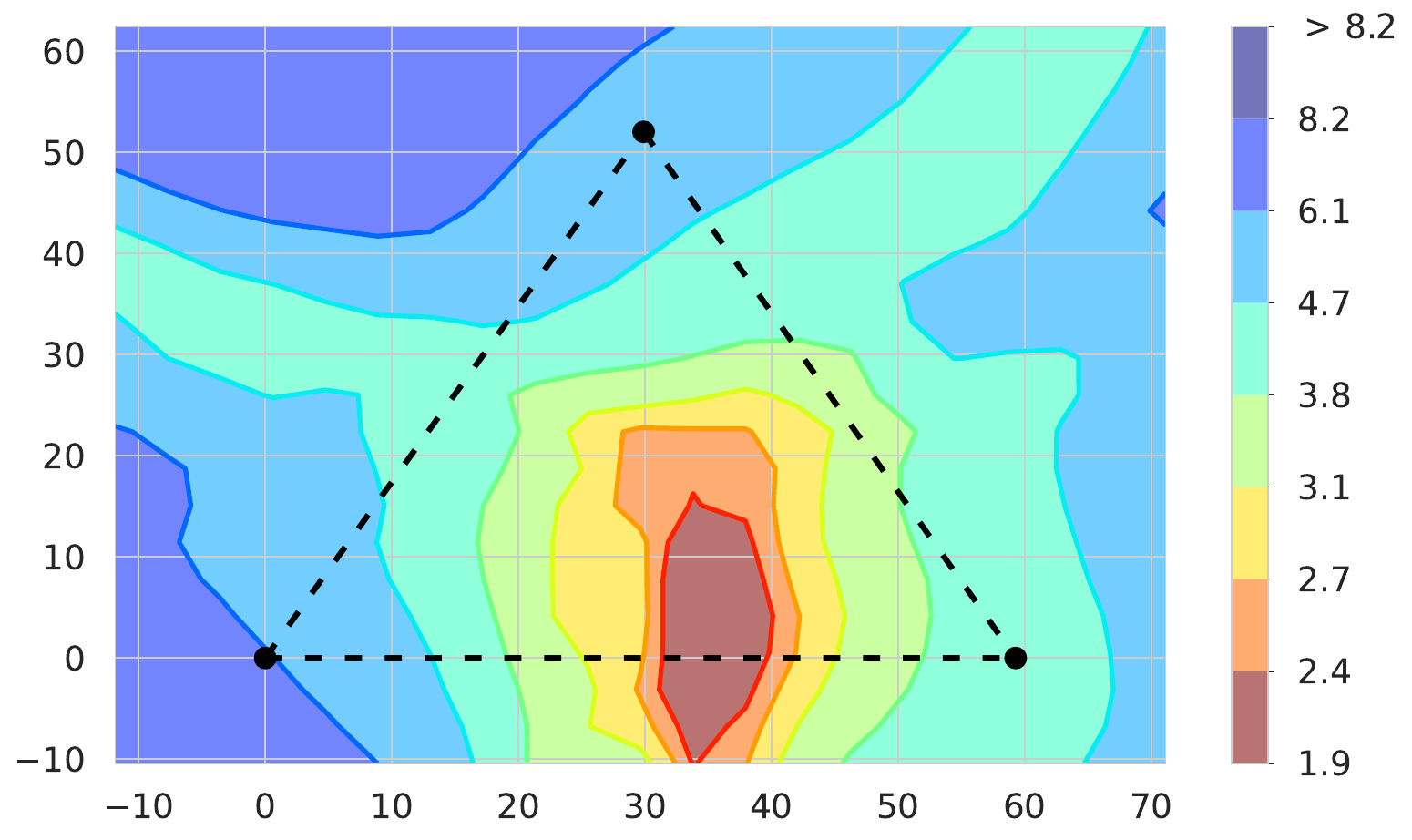}}
    
    \caption{An illustration on the loss landscape of InD and OoD data, as a function of network weights in a two-dimensional subspace.
    The 3 isolated modes are located in similar low loss regions on the InD data, but in significantly different loss areas for the OoD data. The visualization technique follows \cite{garipov2018loss}. Refer to Fig.\ref{fig:method-loss-landscape} for more details on each mode.}
    \label{fig:intro-loss-landscape}
\end{figure*}

In the existing works for OoD detection, the key is to find the different responses from  DNNs towards InD and OoD data and then to leverage such differences to distinguish OoD samples. 
There are various ways to measure such differences with DNNs, such as the logits or probabilities \cite{hendrycks2016baseline,liang2018enhancing,liu2020energy,wang2022vim}, the features or activations \cite{lee2018simple,sun2021react,song2022rankfeat,zhu2022boosting,sun2022out,wang2022vim,yu2023block},  the parameter gradients \cite{lee2020gradients,huang2021importance}, etc.
For example, an early work MSP \cite{hendrycks2016baseline} directly uses the maximum softmax probability to measure the abnormal predictive confidence from DNNs on OoD data.
GradNorm \cite{huang2021importance}, as its name suggests, uses the norm of the gradients as the detection metric, since the gradients of InD and OoD data are distinctively different \textit{w.r.t} the carefully-designed loss in the paper.
A recent work RankFeat \cite{song2022rankfeat} proposes to remove the rank-1 feature matrix in the forward propagation during inference, as the rank-1 matrix is the mostly likely to contain information causing the over-confidence of OoD predictions.

In this work, we address the OoD detection problem from a novel perspective: {\it loss landscape} and {\it mode ensemble}.
The loss landscape of DNNs is extremely complex and highly non-convex due to the millions of parameters and non-linear activations, which inevitably results in numerous isolated optima (\textit{modes}) in the loss landscape, suggested by lots of pioneering works \cite{draxler2018essentially,garipov2018loss,fort2019deep,wortsman2021learning}.
Nevertheless, these works all focus on mode properties in the loss landscape of InD data.
By introducing the loss landscape of OoD data and visualizing locations of 3 isolated modes in Fig.\ref{fig:intro-loss-landscape} and Fig.\ref{fig:method-loss-landscape}, we observe a key phenomenon on those isolated modes that are trained independently on InD data {\it w.r.t} different random seeds:
These modes, even though located in isolation in the loss landscape, undoubtedly all reach low-loss regions on the InD data, shown in Fig.\ref{fig:intro-loss-landscape-1} and Fig.\ref{fig:intro-loss-landscape-2}, yet, their locations in the loss landscape for the OoD data instead hold strong diversities, \textit{i.e.}, these modes show significantly distinct loss regions for the OoD data, shown in Fig.\ref{fig:intro-loss-landscape-3} and Fig.\ref{fig:method-loss-landscape} with more details.

\begin{figure}[ht]
\centering
\includegraphics[width=0.5\linewidth]{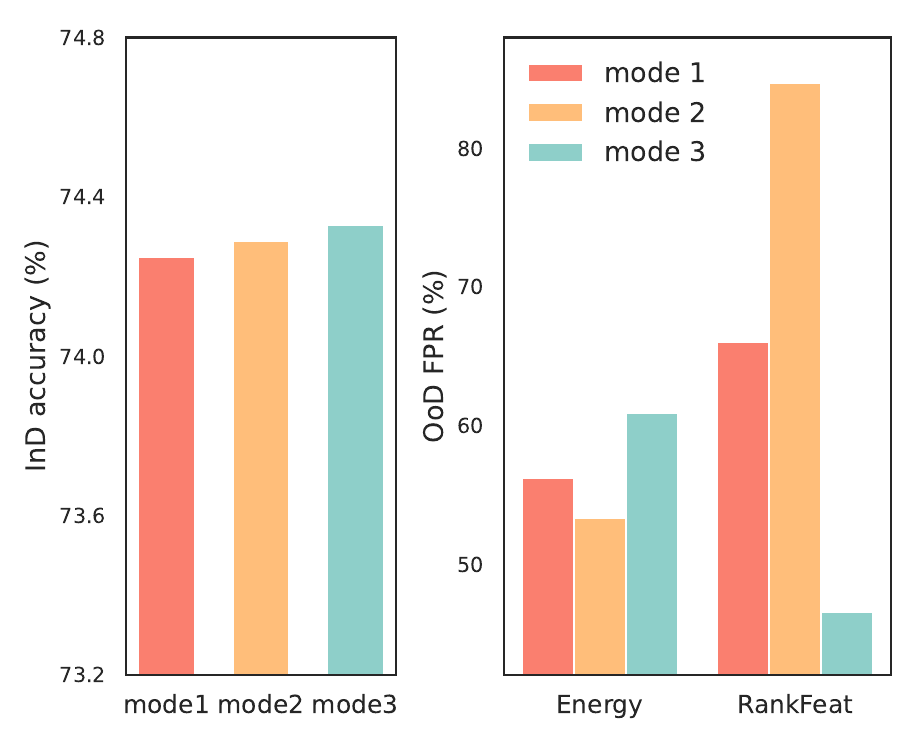}
\caption{There exists a high variance among the OoD detection results (FPR95) of 3 independent modes on 2 OoD detectors Energy \cite{liu2020energy} and RankFeat \cite{song2022rankfeat} (the right panel), while the 3 modes all hold good and similar recognition accuracy on the InD test set (the left panel).
}
\label{fig:intro-var}
\end{figure}

Considering such diversities of these modes in the OoD loss landscape, there naturally exists a hypothesis that these independent modes, though all with consistently good performance on InD data, might hold a large variance on the OoD detection results even under the same OoD detector.
For example, in Fig.\ref{fig:intro-var}, the RankFeat method \cite{song2022rankfeat} achieves fluctuating FPR values ranging from 46.58\% to 84.70\% among isolated modes of the same network structure on the same OoD data set.
Besides, detection results between 2 OoD detectors could be opposite on isolated modes, {\it e.g.}, RankFeat outperforms Energy \cite{liu2020energy} on mode-3 but falls inferior on mode-1 and mode-2.
Such high detection variances among independent modes have been ignored by the research community and bring difficulties in developing and evaluating OoD detectors.
All these findings hence motivate us to explore OoD detection across independent modes, rather than within a single mode, to leverage the diversities reflected on the loss landscape of OoD data.

Accordingly, we revisit the deep ensemble method \cite{lakshminarayanan2017simple} to perform mode ensemble for OoD detection, which is shown to bring substantial variance reduction and performance improvements, and helps the evaluation on OoD detectors.
Specifically, for different types of OoD detectors, we design corresponding mode ensemble strategies by ensembling the output logits or features from multiple isolated modes, then the ensembled outputs are exploited by those OoD detectors.
Besides, a theoretical derivation is provided to show the superiority of ensembling modes over the average of single modes.
Our experiments for mode ensemble are extensive and comprehensive: (i) Various state-of-the-art (SOTA) OoD detection methods are involved, including logits-, features- and gradients-based detectors; (ii) Small-scale, large-scale data sets and distinct network structures, such as the prevailing Vision Transformer (ViT, \cite{dosovitskiy2020image}), are covered. 
All these theoretical and empirical results support our viewpoint: there exist high variances in OoD detection performance of independent modes, and mode ensemble could bring substantial performance improvements and help the evaluation of OoD detectors, which we hope could benefit the research community with the perspective of independent modes in the loss landscape of OoD data.

Our contributions are summarized as follows:
\begin{itemize}
    \item We propose a novel perspective, \textit{loss landscape} and \textit{mode ensemble}, to analyze and address the OoD detection problem. 
    Our observation shows that isolated modes hold diversified loss landscapes on OoD data, resulting in high variances of OoD detection results among these modes.
    \item We revisit the deep ensemble method to perform mode ensemble, design corresponding ensemble strategies for different types of OoD detectors, and find more stable and better OoD detection results.
    \item Extensive and comprehensive empirical results including ablation studies and comparisons with other ensemble methods have verified the uncertainties of single modes and advantages of mode ensemble in the OoD detection task.
\end{itemize}
In the following, Sec.\ref{sec:ood-ensemble} elaborates our mode ensemble for OoD detection.
Experiments are presented in Sec.\ref{sec:experiment} and related works are shown in Sec.\ref{sec:related-work}. Section \ref{sec:conclusion} draws conclusions and discussions.

\section{OoD detection with ensemble modes}
\label{sec:ood-ensemble}
In this section, we firstly outline the background of OoD detection in Sec.\ref{sec:preliminary-ood}, then give detailed elaboration on the loss landscape and mode ensemble of our method in Sec.\ref{sec:mode-ensemble}, and provide theoretical analyses in Sec.\ref{sec:theory}.

\subsection{Preliminary: OoD detection}
\label{sec:preliminary-ood}
Given a neural network $f:\mathbb{R}^D\rightarrow\mathbb{R}^C$, its training data is from some distribution $\mathcal{P}_{\rm in}$, known as the in distribution.
Once deployed, $f$ will encounter unseen data that could be from either the known distribution $\mathcal{P}_{\rm in}$ or an unknown out distribution ${\cal P}_{\rm out}$.
OoD detection aims at deciding whether one new sample $\boldsymbol{x}\in\mathbb{R}^D$ is from ${\cal P}_{\rm in}$ or ${\cal P}_{\rm out}$.

Generally, OoD detection is formulated as a binary classification problem with a decision function $D(\cdot)$ and a scoring function $S(\cdot)$:
\begin{equation}
\label{eq:ood-scoring}
D(\boldsymbol{x})=
\left\{
\begin{array}{ll}
     \mathrm{InD},& S(\boldsymbol{x})>\lambda,\\
     \mathrm{OoD},&S(\boldsymbol{x})<\lambda.
\end{array}
\right.
\end{equation}
The scoring function $S(\cdot)$ assigns a score $S(\boldsymbol{x})$ for a new sample $\boldsymbol{x}$.
If the score $S(\boldsymbol{x})$ is greater than a threshold $\lambda$, then $\boldsymbol{x}$ will be viewed as an InD sample by the decision function $D(\cdot)$, and vice versa.
$\lambda$ is appropriately chosen so that a large proportion (\textit{e.g.}, 95\%) of InD data will be correctly classified.
A justified scoring function $S(\cdot)$ is the key to successfully distinguish OoD data from InD data.

\subsection{Mode ensemble for OoD detection}
\label{sec:mode-ensemble}

\subsubsection{Loss landscape}

\begin{figure*}[ht]
\centering
\includegraphics[width=0.9\linewidth]{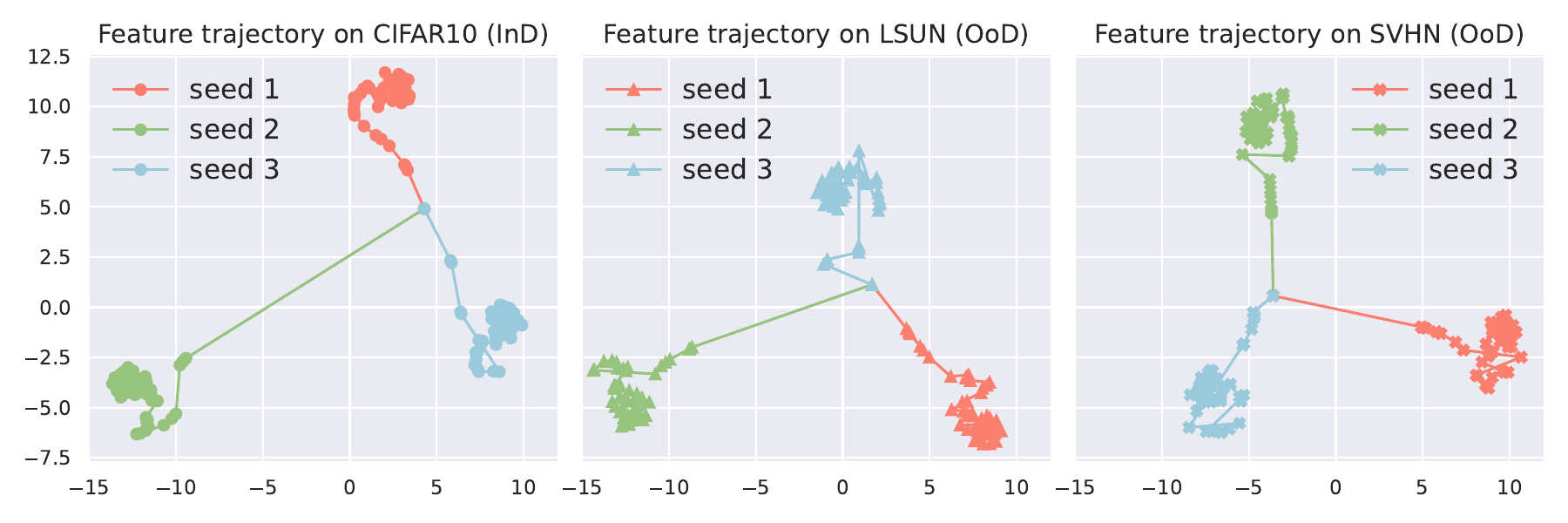}
\caption{An illustration on the feature trajectories during model training on CIFAR10 (InD) {\it w.r.t} 3 random seeds. 
48 checkpoints in each training are sampled.
The learned features of CIFAR10 (left), LSUN (OoD, middle) and SVHN (OoD, right) by these checkpoints are reduced to 2-dimension via t-SNE \cite{van2008visualizing}, showing clearly the resulting 3 isolated modes.
}
\label{fig:feat-traj}
\end{figure*}

Inspired by prior works investigating the loss landscape and mode connectivity for the generalization ability of DNNs on the InD test data \cite{draxler2018essentially,garipov2018loss,fort2019deep,wortsman2021learning}, we propose to analyze the OoD detection problem from the loss landscape perspective.
Specifically, our analysis focuses on those {\it independent} or {\it isolated} modes that are obtained by training multiple DNNs merely on the InD data \textit{w.r.t} different random seeds.
As the random seed dominates the initialization of DNNs and the noise in the SGD optimization, these modes are actually in correspondence to unique training trajectories, and thus are independent and isolated.
Fig.\ref{fig:feat-traj} illustrates 3 distinct training trajectories of 3 modes \textit{w.r.t} 3 different random seeds by visualizing features learned by the checkpoints during training.

\begin{figure*}[t]
    \centering
    \subfigure[InD data, test set, \textbf{mode-1}.]{\includegraphics[width=0.32\linewidth]{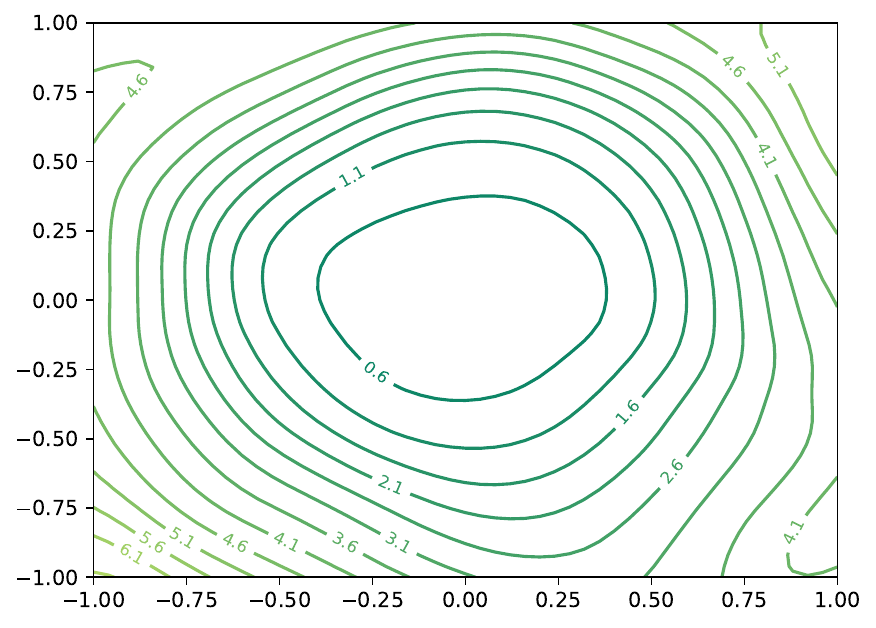}}
    \subfigure[InD data, test set, \textbf{mode-2}.]{\includegraphics[width=0.32\linewidth]{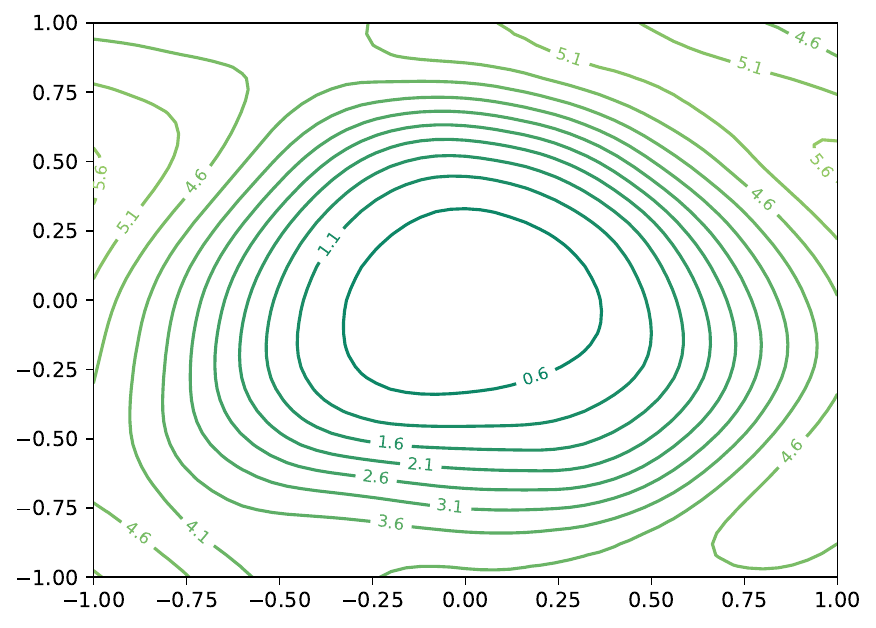}}
    \subfigure[InD data, test set, \textbf{mode-3}.]{\includegraphics[width=0.32\linewidth]{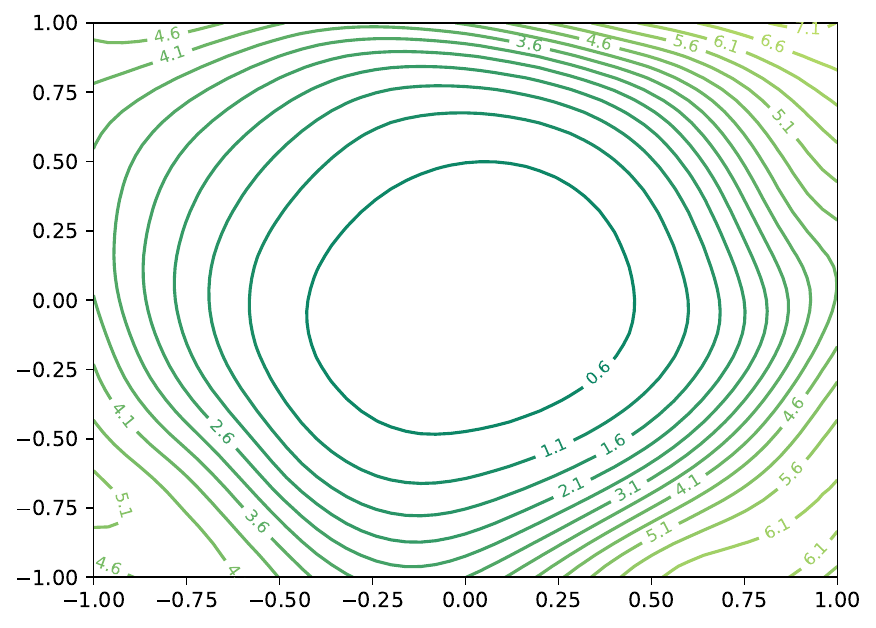}}
    \subfigure[OoD data set, \textbf{mode-1}.]{\includegraphics[width=0.32\linewidth]{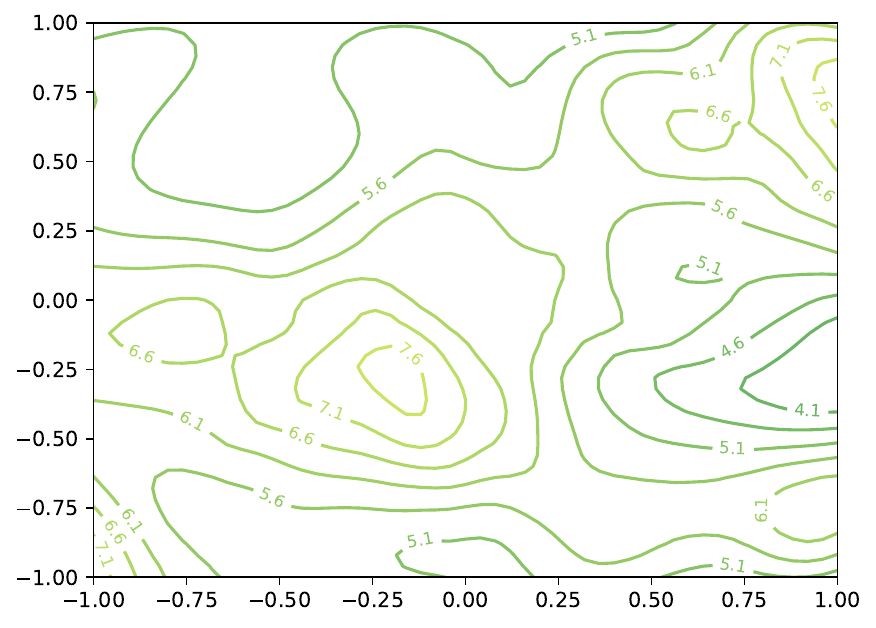}}
    \subfigure[OoD data set, \textbf{mode-2}.]{\includegraphics[width=0.32\linewidth]{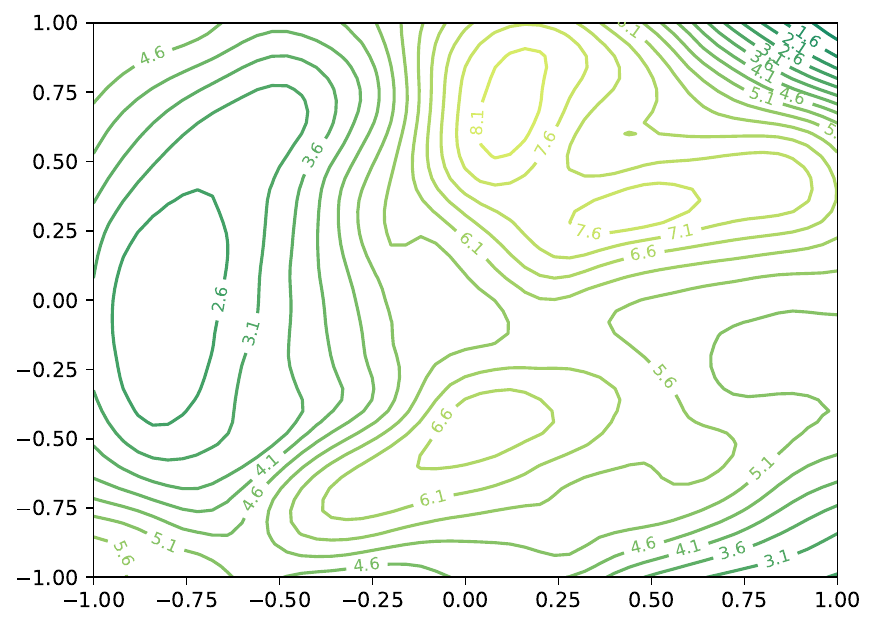}}
    \subfigure[OoD data set, \textbf{mode-3}.]{\includegraphics[width=0.32\linewidth]{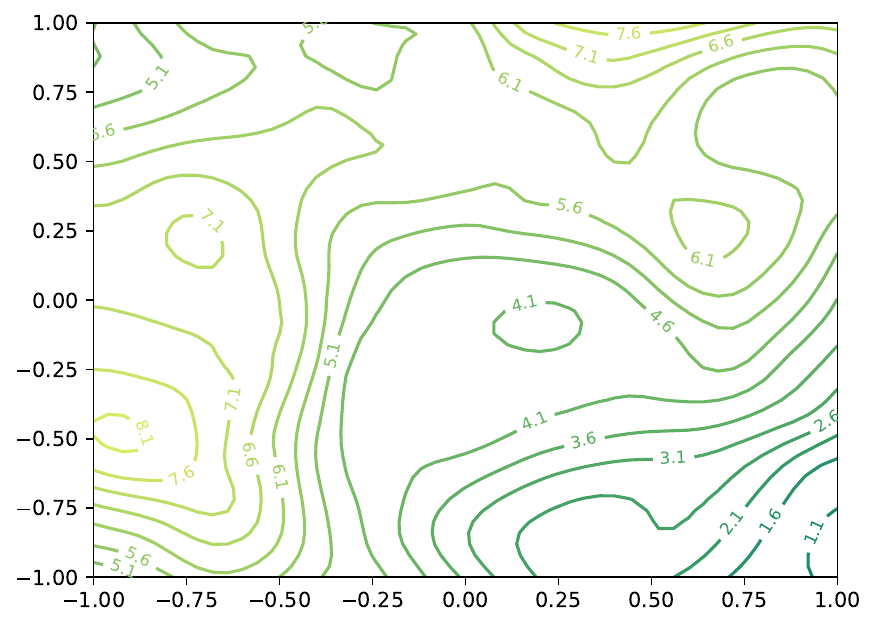}}
    
    \caption{An illustration of the loss landscapes of 3 independent modes (left, middle and right, respectively) on the InD (top) and OoD (bottom) data.
    The visualization technique follows \cite{li2018visualizing}.}
    \label{fig:method-loss-landscape}
\end{figure*}

As these modes are well-trained on the InD data, though they are independent and isolated with each other, their locations in the loss landscape of InD data still share similarities, \textit{i.e.}, low-loss regions of the InD data.
However, for the unseen data from an \textit{out} distribution, these modes might be located in distinct positions with either high or low losses on the OoD data.
In Fig.\ref{fig:method-loss-landscape}, as a complement to Fig.\ref{fig:intro-loss-landscape}, the loss landscapes of each mode on the InD and OoD data are visualized, and there clearly exist notable differences in the OoD loss landscapes among those independent modes.

Accordingly, the diversities on the loss landscapes (Fig.\ref{fig:method-loss-landscape}) and features (Fig.\ref{fig:feat-traj}) among the independent modes inspire us to investigate the OoD detection problem across such modes.
As the existing OoD detectors mainly focus on one single mode and rely on the differences in the logits or features between InD and OoD data, a hypothesis naturally comes to the mind:

{\centering\textit{The OoD detection performance of one single mode is of high uncertainty, since such a mode, well-trained on InD data, is possibly located in diverse positions of significantly different loss landscapes of OoD data.}}

Our experiments in Sec.\ref{sec:experiment} and Appendix \ref{app-sec:all-exp} further validate this hypothesis, {\it i.e.}, the high variances of the OoD detection performance across independent modes.

Such a phenomenon has not been explored by the research community.
The high variances among independent modes hinder the development and evaluation of OoD detectors.
In this regard, we hope our work could benefit OoD detection researches from the perspective of independent modes in the loss landscape, as an alternative aside from exploiting the outputs of single DNNs.

\subsubsection{Mode ensemble} 
To further exploit such diversities on the OoD loss landscape and to reduce the high variances across modes, we revisit the deep ensemble method \cite{lakshminarayanan2017simple}, leverage multiple independent modes to perform mode ensemble, and find significant variance reduction with substantial performance improvements in OoD detection.

Nevertheless, how to efficiently involve various OoD detection methods into the mode ensemble framework remains a troublesome issue.
On the one hand, as pointed in \cite{xue2022boosting}, directly ensembling the ``yes or no'' detection results from multiple single modes will accumulate the misclassification probabilities.
On the other hand, current detectors work on different outputs of DNNs, including logits, features and gradients, which indicates that the mode ensemble should be carefully designed for specific detectors.

Given $N$ modes $f_{s_1},\cdots,f_{s_N}:\mathbb{R}^D\rightarrow\mathbb{R}^C$ trained independently \textit{w.r.t} $N$ random seeds $s_1,\cdots,s_N$, in this work, the mode ensemble during inference takes a general form as:
\begin{equation}
\label{eq:ensemble}
\boldsymbol{h}_{\rm ens}=\frac{1}{N}\sum_{i=1}^N\boldsymbol{h}_{s_i},
\end{equation}
where $\boldsymbol{h}_{s_i}$ denotes the output features or logits from $f_{s_i}$. 
Such ensembled features or logits $\boldsymbol{h}_{\rm ens}$ from multiple isolated modes are then exploited by different OoD detectors to achieve mode ensemble for OoD detection.
We put complete and thorough details of the mode ensemble for representative OoD detectors in Appendix \ref{app-sec:ood-baseline-ensemble}.
In this work, we mainly focus on those \textit{post-hoc} and \textit{OoD-agnostic} detectors, covering all aforementioned 3 types, since such detectors do not require OoD data nor additional model modifications during training, and execute detection strategies only in the inference stage.

\paragraph{Discussions with other ensemble methods for OoD detection \cite{vyas2018out,yang2021ensemble,xue2022boosting}}
Among these works, \cite{vyas2018out} requires training DNNs with a margin loss involving additional OoD data, which is orthogonal to our settings that models trained solely on InD data are considered.
\cite{yang2021ensemble} introduces the contrastive loss and the triplet loss to train Siamese networks and triplet networks for ensemble, where the trained networks are not independent with each other.
In \cite{xue2022boosting}, a novel p-value-based ensemble strategy is proposed for detecting OoD samples, which differs from the common threshold-based decision rule.
Our loss landscape perspective motivates us to focus on the general case: standard models trained without OoD data, in correspondence to those post-hoc, OoD-agnostic and model-agnostic OoD detectors.

\subsection{Theoretical analysis}
\label{sec:theory}
In this section, we provide a theoretical discussion on how ensemble boosts OoD detection performance.
Our analysis follows the setups in \cite{miller2021accuracy} and shows the superiority of ensembling modes over the average of single modes in terms of accuracy difference between InD data and OoD data.

Let us consider a binary classification problem and suppose the in distribution ${\cal P}_{\rm in}$ is an isotropic Gaussian:
\begin{equation}
\label{eq:p_in}
\mathcal{P}_{\rm in}=\{(\boldsymbol{x},y)\ |\ \boldsymbol{x}\sim\mathcal{N}(\boldsymbol{\mu}\cdot y;\sigma^2\boldsymbol{I}_{D\times D})\},
\end{equation}
with $\boldsymbol{x}\in\mathbb{R}^D$, $y\in\{-1,1\}$, the mean vector $\boldsymbol{\mu}\in\mathbb{R}^D$ and the variance $\sigma^2$ ($\sigma>0$).
The out distribution ${\cal P}_{\rm out}$ is defined as a shift from $\mathcal{P}_{\rm in}$:
\begin{equation}
\begin{aligned}
\label{eq:p_out}
\mathcal{P}_{\rm out}=\{(\boldsymbol{x},y)\ |\ \boldsymbol{x}\sim\mathcal{N}(\boldsymbol{\mu}^\prime\cdot y;{\sigma^\prime}^2\boldsymbol{I}_{D\times D}),\quad\boldsymbol{\mu}^\prime=\alpha\cdot\boldsymbol{\mu}+\beta\cdot\boldsymbol{\Delta},\sigma^\prime=\gamma\cdot\sigma\},
\end{aligned}
\end{equation}
where $\alpha,\beta,\gamma>0$ are fixed scalars and $\boldsymbol{\Delta}\in\mathbb{R}^D$ is an unknown random distribution.

For mode ensemble, we consider $N$ linear classifiers in the form of $g_i:\boldsymbol{x}\rightarrow\mathrm{sign}(\boldsymbol{\theta}_i^\top\boldsymbol{x}),i=1,\cdots,N$, learned \textit{w.r.t} $N$ different random seeds on $\cal{P}_{\rm in}$, each with weights $\boldsymbol{\theta}_i\in\mathbb{R}^D$.
The mode ensemble function is thus defined as $g_{\rm ens}:\boldsymbol{x}\rightarrow\mathrm{sign}(\sum_{i=1}^{N}(\boldsymbol{\theta}^\top_i\boldsymbol{x}))$.
The following proposition states the superiority of mode ensemble in achieving a lower gap on the accuracy between InD data and OoD data over the average of multiple modes.

\begin{proposition}
\label{prop:gap}
Consider the in distribution ${\cal P}_{\rm in}$, out distribution ${\cal P}_{\rm out}$, and $N$ independent modes $g_i, i=1,\cdots,N$ defined above, we have $\mathcal{G}(g_{\rm ens},\mathcal{P}_{\rm in},\mathcal{P}_{\rm out})\leq\frac{1}{N}\sum_{i=1}^N\mathcal{G}(g_i,\mathcal{P}_{\rm in},\mathcal{P}_{\rm out})$, where
\begin{equation}
\begin{aligned}
\label{eq:acc-gap}
\mathcal{G}(g,\mathcal{P}_{\rm in},\mathcal{P}_{\rm out})=\left|\Phi^{-1}(\mathcal{ACC}(g,\mathcal{P}_{\rm out}))-\frac{\alpha}{\gamma}\Phi^{-1}(\mathcal{ACC}(g,\mathcal{P}_{\rm in}))\right|
\end{aligned}
\end{equation}
denotes the gap of the probit-transformed accuracy between $\mathcal{P}_{\rm in}$ and $\mathcal{P}_{\rm out}$ achieved by the classifier $g$.
$\Phi^{-1}$ indicates the inverse of the standard normal cumulative distribution function $\Phi(x)=\int_{-\infty}^x\frac{1}{\sqrt{2\pi}}e^{-t^2/2}{\rm d}t$.
\end{proposition}
The proof of this proposition is deferred to Appendix \ref{app-sec:theory-proof}.
In the next section, extensive empirical results will verify the high variances on the OoD detection performance among independently trained models on InD data, and the substantial improvements brought by mode ensemble.

\section{Experiments}
\label{sec:experiment}
In experiments, we will empirically show (i) the high variances of independent modes on the OoD detection performance in Sec.\ref{sec:exp-single-modes}, (ii) the substantial improved detection results via mode ensemble in Sec.\ref{sec:exp-mode-ensemble}, and (iii) more in-depth empirical discussions on the mode ensemble in Sec.\ref{sec:exp-in-depth}.

\subsection{Setups}
\label{sec:exp-setups}
Following previous works on OoD detection \cite{hendrycks2016baseline,liang2018enhancing,lee2018simple,liu2020energy,sun2021react,huang2021importance,song2022rankfeat,zhu2022boosting,sun2022out,wang2022vim,yu2023block}, our experiments are executed on the small-scale CIFAR10 \cite{krizhevsky2009learning} and the large-scale ImageNet-1K \cite{deng2009imagenet} benchmarks.
To obtain multiple isolated modes, for the former, 10 models are independently trained on CIFAR10 \textit{w.r.t} 10 different random seeds, including structures of ResNet18 \cite{he2016deep} and Wide ResNet28x10 \cite{zagoruyko2016wide}.
For the latter, 5 models of ResNet50 \cite{he2016deep} and DenseNet121 \cite{huang2017densely} are independently trained \textit{w.r.t} 5 different random seeds on ImageNet-1K.
We further train 3 independent models of T2T-ViT-14 \cite{yuan2021tokens} \textit{w.r.t} 3 different random seeds from scratch on ImageNet-1K to further evaluate the mode performance on such a transformer structure.

For CIFAR10 as the InD data, we evaluate the OoD detection performance of different modes on 5 OoD data sets: SVHN \cite{netzer2011reading}, LSUN \cite{yu2015lsun}, iSUN \cite{xu2015turkergaze}, Textures \cite{cimpoi2014describing} and Places \cite{zhou2017places}.
For ImageNet-1K as the InD data, the selected OoD data sets include iNaturalist \cite{van2018inaturalist}, SUN \cite{xiao2010sun}, Places \cite{zhou2017places} and Textures \cite{cimpoi2014describing}.
Refer to Appendix \ref{app-sec:exp-setup-details-datasets} for details of these data sets.

We select 7 representative and SOTA OoD detectors as the baselines, including MSP \cite{hendrycks2016baseline}, ODIN \cite{liang2018enhancing}, Energy \cite{liu2020energy},  Mahalanobis \cite{lee2018simple}, KNN \cite{sun2022out}, RankFeat \cite{song2022rankfeat} and GradNorm \cite{huang2021importance}.
These are all post-hoc and OoD-agnostic detectors, covering logits-based (MSP, ODIN, Energy), features-based (Mahalanobis, KNN and RankFeat) and gradients-based (GradNorm) types.
We evaluate the detection performance of these methods on multiple independent modes to show the high detection variances existing widely in different types of detectors and network structures, and then perform mode ensemble to reduce the variances and boost OoD detection performance.
For each single and ensembling modes, we record their (i) InD classification accuracy, (ii) false positive rate (FPR) on OoD data sets when the true positive rate on InD data is 95\% and (iii) area under the receiver operating characteristic curve (AUROC).

\noindent\textbf{Remark}
For checkable reproducibility, all the training and evaluation code and the trained checkpoints are released\footnote{\href{https://github.com/fanghenshaometeor/ood-mode-ensemble}{https://github.com/fanghenshaometeor/ood-mode-ensemble}} publicly.
The details of model training can also be found in Appendix \ref{app-sec:exp-setup-details-training}.
Besides, for a clear elaboration, in the following Sec.\ref{sec:exp-single-modes} and Sec.\ref{sec:exp-mode-ensemble}, we mainly present results of DenseNet121 and ViT on ImageNet-1K and the 6 common detectors, including 3 logits-based (MSP, ODIN and Energy), 1 gradients-based (GradNorm), and 2 features-based (Mahalanobis and RankFeat) detectors.
Comprehensive results covering more modes on other InD and OoD datasets, baseline detectors and network structures are provided in Appendix \ref{app-sec:all-exp}.

\begin{table}[t]
    \centering
    \begin{tabular}{@{}c c c c c c@{}}
    \toprule
    mode & mode-1 & mode-2 & mode-3 & mode-4 & mode-5 \\
    \midrule
    InD Acc. & 74.25 & 74.30 & 74.46 & 74.29 & 74.33 \\
    \bottomrule
    \end{tabular}
    \caption{The classification accuracy on the ImageNet-1K validation set of 5 isolated modes of DenseNet121.}
    \label{tab:exp-indacc-imgnet}
\end{table}

\subsection{High variances among the independent modes for OoD detection}
\label{sec:exp-single-modes}

This subsection empirically presents the existence of a high variance in the OoD detection results of multiple isolated modes.
The classification accuracy of these modes on the InD data is shown in Tab.\ref{tab:exp-indacc-imgnet} to illustrate that these modes are all well-trained on InD data.
Then, regarding the OoD detection, the detection results of each mode \textit{w.r.t} different detectors are reported in Tab.\ref{tab:exp-uncertainty-imgnet}.

\begin{table*}[t]
    \centering
    \begin{tabular}{@{}c H cc cc cc cc HH@{}}
    \toprule
    \multirow{3}{*}{Modes} & \multirow{3}{*}{InD ACC.}  & \multicolumn{10}{c}{OoD data sets} \\
    & & \multicolumn{2}{c}{iNaturalist} & \multicolumn{2}{c}{SUN} & \multicolumn{2}{c}{Places} & \multicolumn{2}{c}{Textures} & & \\
    & & FPR$\downarrow$ & AUROC$\uparrow$ & FPR$\downarrow$ & AUROC$\uparrow$ &
    FPR$\downarrow$ & AUROC$\uparrow$ & FPR$\downarrow$ & AUROC$\uparrow$ & FPR$\downarrow$ & AUROC$\uparrow$ \\
    
    \midrule
    \multicolumn{12}{c}{\textbf{MSP}}\\
    mode-1 & & 58.44 & 87.18 & \uwave{\bf 71.11} & 80.78 & \uwave{\bf 74.11} & \uwave{\bf 79.69} & 67.07 & 80.72 \\
    mode-2 & & \underline{\bf 55.68} & \underline{\bf 87.28} & \underline{\bf 68.78} & \underline{\bf 81.49} & \underline{\bf 70.85} & \underline{\bf 80.63} & 66.21 & \uwave{\bf 80.27} \\
    mode-3 & & 58.29 & 87.10 & 70.60 & 81.02 & 73.01 & 80.07 & 66.06 & 80.70 \\
    mode-4 & & 58.52 & 87.20 & 70.73 & 80.81 & 72.79 & 79.77 & \uwave{\bf 67.85} & 80.46 \\
    mode-5 & & \uwave{\bf 60.97} & \uwave{\bf 86.19} & 70.82 & \uwave{\bf 80.69} & 72.76 & 79.89 & \underline{\bf 64.41} & \underline{\bf 80.93} \\

    \midrule
    \multicolumn{12}{c}{\textbf{ODIN}}\\
    mode-1 & & 54.88 & 89.39 & \uwave{\bf 60.26} & \uwave{\bf 85.76} & \uwave{\bf 65.84} & \uwave{\bf 83.33} & 54.36 & 86.61 \\
    mode-2 & & \underline{\bf 50.56} & 90.21 & \underline{\bf 55.79} & \underline{\bf 87.00} & \underline{\bf 60.81} & \underline{\bf 85.08} & 53.35 & \underline{\bf 86.62} \\
    mode-3 & & 51.61 & 90.12 & 57.90 & 86.08 & 63.82 & \underline{\bf 85.08} & 53.35 & \underline{\bf 86.62} \\
    mode-4 & & 51.47 & \underline{\bf 90.26} & 59.03 & 86.28 & 63.56 & 84.01 & \uwave{\bf 55.12} & 86.20 \\
    mode-5 & & \uwave{\bf 58.13} & \uwave{\bf 88.39} & 59.17 & 85.84 & 63.17 & 84.07 & \underline{\bf 51.83} & \uwave{\bf 86.19} \\

    \midrule
    \multicolumn{12}{c}{\textbf{Energy}}\\
    mode-1 & & 56.24 & 88.63 & \uwave{\bf 57.85} & \uwave{\bf 85.99} & \uwave{\bf 64.35} & \uwave{\bf 83.29} & 51.93 & 86.98 \\
    mode-2 & & \underline{\bf 52.86} & 89.67 & \underline{\bf 53.31} & \underline{\bf 87.32} & \underline{\bf 59.77} & \underline{\bf 85.16} & 50.76 & \underline{\bf 87.05} \\
    mode-3 & & 53.19 & 89.61 & 55.89 & 86.35 & 62.05 & 84.13 & 50.64 & 86.86 \\
    mode-4 & & 53.31 & \underline{\bf 89.77} & 56.85 & 86.58 & 62.75 & 84.08 & \uwave{\bf 53.56} & 86.53 \\
    mode-5 & & \uwave{\bf 60.92} & \uwave{\bf 87.56} & 56.76 & 86.07 & 61.93 & 84.11 & \underline{\bf 49.93} & \uwave{\bf 86.46} \\ 
    
    \midrule
    \multicolumn{12}{c}{\textbf{GradNorm}}\\
    mode-1 & & 73.79 & 77.93 & 72.64 & 75.03 & \uwave{\bf 81.21} & 70.23 & 66.15 & 77.82 \\
    mode-2 & & 69.99 & 79.89 & \underline{\bf 68.30} & 76.72 & \underline{\bf 75.71} & 72.63 & \underline{\bf 63.30} & \underline{\bf 78.74} \\
    mode-3 & & 69.93 & 79.67 & 69.12 & \underline{\bf 77.56} & 77.22 & \underline{\bf 73.34} & 64.06 & 78.48 \\
    mode-4 & & \underline{\bf 66.85} & \underline{\bf 81.07} & 72.51 & 75.61 & 80.08 & 71.44 & 65.87 & 77.27 \\
    mode-5 & & \uwave{\bf 77.07} & \uwave{\bf 73.47} & \uwave{\bf 73.66} & \uwave{\bf 73.88} & 80.59 & \uwave{\bf 69.90} & \uwave{\bf 66.83} & \uwave{\bf 76.39} \\   

    \midrule
    \multicolumn{12}{c}{\textbf{Mahalanobis}}\\
    mode-1 & & 89.20 & 63.79 & 84.50 & \underline{\bf70.98} & 83.84 & 71.65 & \underline{\bf43.56}	& \underline{\bf75.46} \\
    mode-2 & & \uwave{\bf93.31}	& 60.90 & 85.98 &	68.14 &	83.87 &	71.11 &	\uwave{\bf65.21} &	\uwave{\bf58.26}\\
    mode-3 & & \underline{\bf88.43} & \underline{\bf69.11} & 86.21 & 70.74 &	85.17 &	72.50 &	51.40 &	70.16\\
    mode-4 & & 91.84 & \uwave{\bf58.88} & \uwave{\bf86.82} & \uwave{\bf66.23} &	\uwave{\bf85.54} &	\uwave{\bf69.64} &	49.86 &	68.64\\
    mode-5 & & 92.43 & 59.27 & \underline{\bf82.07} & 70.24 & \underline{\bf81.02} &	\underline{\bf72.95} &	57.20 & 63.21\\

    \midrule
    \multicolumn{12}{c}{\textbf{RankFeat}}\\
    mode-1 & & 66.01 & 85.91 & \uwave{\bf 75.53} & \uwave{\bf 80.27} & \uwave{\bf 79.95} & \uwave{\bf 75.64} & 43.60 & \uwave{\bf 90.35} \\
    mode-2 & & 58.49 & 83.77 & \underline{\bf 34.70} & \underline{\bf 92.02} & \underline{\bf 50.70} & \underline{\bf 86.12} & 32.73 & 92.45 \\
    mode-3 & & 59.53 & 85.17 & 50.07 & 88.37 & 63.27 & 82.05 & 40.64 & 91.54 \\
    mode-4 & & \uwave{\bf 84.70} & \uwave{\bf 78.61} & 69.57 & 83.82 & 76.45 & 78.21 & \uwave{\bf 49.89} & 90.81 \\
    mode-5 & & \underline{\bf 46.58} & \underline{\bf 87.27} & 44.46 & 88.42 & 58.95 & 81.54 & \underline{\bf 22.48} & \underline{\bf 94.15} \\
    
    \bottomrule
    \end{tabular}
    \caption{The detection performance of each independent mode (DenseNet121 trained on ImageNet-1K) on each OoD data set \textit{w.r.t} different types of OoD detectors.
    The \underline{\bf best} and \uwave{\bf worst} of the detection results among the 5 modes are highlighted.}
    \label{tab:exp-uncertainty-imgnet}
\end{table*}

In Tab.\ref{tab:exp-uncertainty-imgnet}, we mark the \underline{\bf best} and \uwave{\bf worst} FPR and AUROC results of the 5 modes on each OoD data set achieved by each OoD detector.
Though these modes all perform consistently well on the InD data in Tab.\ref{tab:exp-indacc-imgnet}, there clearly exists a large variance of the OoD detection results across independent modes for different detectors.
Specifically, we can find the following two phenomena.
\begin{itemize}
    \item For the same OoD detector, independent modes hold significantly-fluctuating FPR results.
    For example, the RankFeat \cite{song2022rankfeat} even reaches a worst FPR value 84.71\% by the mode-4 and a best FPR value 46.58\% by the mode-5 on the iNaturalist \cite{van2018inaturalist} data set.
    \item For different modes, detection results among those methods might be opposite.
    For example, considering the mode-4, Energy \cite{liu2020energy} achieves better detection result on iNaturalist (53.31\%) than RankFeat (84.70\%), while for the mode-5, RankFeat (46.58\%) outperforms Energy (60.92\%) a lot on iNaturalist.
\end{itemize}
Such widely-existed high variances support our viewpoint: \textit{The OoD detection performance of single modes is of high uncertainty}.
In addition, these phenomena are more conspicuous on the small-scale CIFAR10 benchmark with even higher variances of the detection FPR results across independent modes, see Appendix \ref{app-sec:all-exp} for complete results and more analysis.

\subsection{Mode ensemble stabilizes and boosts OoD detection}
\label{sec:exp-mode-ensemble}
In this subsection, mode ensemble is introduced to alleviate the high variances of the OoD detection performance among multiple isolated modes and to further boost the detection.
Table \ref{tab:exp-ensemble-imgnet} shows the OoD detection performance of ensembling $k$ modes.
For $k=1$, we record the mean values and standard deviations of the detection results among the total $N$ modes.
For each $k>1$, firstly $k$ different modes are randomly selected and get ensembled to perform OoD detection, then such process is repeated 3 times to calculate the mean values and standard deviations of the detection results of ensembling $k$ modes.

\begin{table*}[t]
    \centering
    \resizebox{\textwidth}{!}{
    \begin{tabular}{@{}c H cc cc cc cc HH@{}}
    \toprule
    \multirow{3}{*}{\makecell[c]{Ensemble\\of $k$ modes}} & \multirow{3}{*}{InD ACC.}  & \multicolumn{10}{c}{OoD data sets} \\
    & & \multicolumn{2}{c}{iNaturalist} & \multicolumn{2}{c}{SUN} & \multicolumn{2}{c}{Places} & \multicolumn{2}{c}{Textures} & & \\
    & & FPR$\downarrow$ & AUROC$\uparrow$ & FPR$\downarrow$ & AUROC$\uparrow$ &
    FPR$\downarrow$ & AUROC$\uparrow$ & FPR$\downarrow$ & AUROC$\uparrow$ & FPR$\downarrow$ & AUROC$\uparrow$ \\
    \midrule

    \multicolumn{12}{c}{\textbf{MSP}}\\
    $k=1$ & & 58.38$\pm$1.87 & 86.99$\pm$0.45 & 70.41$\pm$0.93 & 80.96$\pm$0.32 & 72.70$\pm$1.17 & 80.01$\pm$0.37 & 66.32$\pm$1.29 & 80.62$\pm$0.26 \\
    $k=2$ & & 54.69$\pm$1.01 & 88.29$\pm$0.24 & 68.75$\pm$0.39 & 82.03$\pm$0.04 & 70.72$\pm$0.37 & 81.21$\pm$0.12 & 64.05$\pm$0.57 & 81.58$\pm$0.05 \\
    $k=3$ & & 53.57$\pm$0.55 & 88.75$\pm$0.19 & 68.31$\pm$0.21 & 82.18$\pm$0.09 & 70.74$\pm$0.37 & 81.38$\pm$0.12 & 63.39$\pm$0.48 & \cellcolor{tabgray}{\bf 81.99$\pm$0.02} \\
    $k=4$ & & \cellcolor{tabgray}{\bf 53.42$\pm$0.49} & \cellcolor{tabgray}{\bf 88.94$\pm$0.11} & \cellcolor{tabgray}{\bf 68.48$\pm$0.15} & \cellcolor{tabgray}{\bf 82.34$\pm$0.01} & \cellcolor{tabgray}{\bf 70.47$\pm$0.35} & \cellcolor{tabgray}{\bf 81.52$\pm$0.06} & \cellcolor{tabgray}{\bf 63.35$\pm$0.78} & 82.11$\pm$0.07 \\
    \midrule

    \multicolumn{12}{c}{\textbf{ODIN}}\\
    $k=1$ & & 53.33$\pm$3.14 & 89.67$\pm$0.80 & 58.43$\pm$1.70 & 86.19$\pm$0.50 & 63.44$\pm$1.80 & 84.31$\pm$0.76 & 53.51$\pm$1.28 & 86.43$\pm$0.22 \\
    $k=2$ & & 50.60$\pm$2.55 & 91.06$\pm$0.56 & 56.50$\pm$1.54 & 87.37$\pm$0.40 & 61.09$\pm$0.96 & 85.47$\pm$0.46 & 49.41$\pm$0.93 & \cellcolor{tabgray}{\bf 88.30$\pm$0.06} \\
    $k=3$ & & 48.43$\pm$0.89 & 91.76$\pm$0.15 & \cellcolor{tabgray}{\bf 55.75$\pm$0.35} & \cellcolor{tabgray}{\bf 87.62$\pm$0.11} & 60.59$\pm$0.49 & \cellcolor{tabgray}{\bf 85.78$\pm$0.16} & \cellcolor{tabgray}{\bf 47.49$\pm$0.22} & 88.86$\pm$0.14 \\
    $k=4$ & & \cellcolor{tabgray}{\bf 48.37$\pm$0.28} & \cellcolor{tabgray}{\bf 91.93$\pm$0.09} & 55.81$\pm$0.50 & 87.74$\pm$0.14 & \cellcolor{tabgray}{\bf 60.62$\pm$0.46} & 85.89$\pm$0.20 & 46.52$\pm$0.57 & 89.13$\pm$0.08 \\
    \midrule
    
    \multicolumn{12}{c}{\textbf{Energy}}\\
    $k=1$ & & 55.30$\pm$3.42 & 89.05$\pm$0.95 & 56.13$\pm$1.72 & 86.46$\pm$0.53 & 62.17$\pm$1.65 & 84.15$\pm$0.66 & 51.36$\pm$1.42 & 86.78$\pm$0.27 \\
    $k=2$ & & 51.33$\pm$1.78 & 90.85$\pm$0.46 & 53.75$\pm$1.14 & 87.64$\pm$0.22 & 59.90$\pm$1.54 & 85.44$\pm$0.48 & 46.00$\pm$1.18 & 88.83$\pm$0.11 \\
    $k=3$ & & 49.58$\pm$1.60 & 91.52$\pm$0.28 & 52.72$\pm$1.04 & 88.05$\pm$0.20 & 58.84$\pm$1.19 & 85.93$\pm$0.33 & 44.27$\pm$0.51 & \cellcolor{tabgray}{\bf 89.48$\pm$0.03} \\
    $k=4$ & & \cellcolor{tabgray}{\bf 49.23$\pm$0.97} & \cellcolor{tabgray}{\bf 91.64$\pm$0.20} & \cellcolor{tabgray}{\bf 52.33$\pm$0.27} & \cellcolor{tabgray}{\bf 88.23$\pm$0.07} & \cellcolor{tabgray}{\bf 58.34$\pm$0.57} & \cellcolor{tabgray}{\bf 86.14$\pm$0.10} & \cellcolor{tabgray}{\bf 42.74$\pm$0.43} & 89.81$\pm$0.08 \\
    \midrule
    
    \multicolumn{12}{c}{\textbf{GradNorm}}\\
    $k=1$ & & 71.53$\pm$3.96 & 78.41$\pm$2.98 & 71.25$\pm$2.38 & 75.76$\pm$1.44 & 78.96$\pm$2.38 & 71.51$\pm$1.49 & 65.24$\pm$1.49 & 77.74$\pm$0.95 \\
    $k=2$ & & 69.97$\pm$3.93 & 79.96$\pm$2.53 & 68.72$\pm$1.53 & 77.06$\pm$0.94 & 77.15$\pm$1.67 & 72.93$\pm$1.08 & 61.93$\pm$0.88 & 79.85$\pm$0.41 \\
    $k=3$ & & 67.58$\pm$2.53 & 81.99$\pm$1.44 & 67.78$\pm$0.41 & 77.97$\pm$0.32 & 76.43$\pm$0.49 & 73.69$\pm$0.27 & 60.71$\pm$0.75 & \cellcolor{tabgray}{\bf 81.05$\pm$0.19} \\
    $k=4$ & & \cellcolor{tabgray}{\bf 67.22$\pm$1.64} & \cellcolor{tabgray}{\bf 82.33$\pm$0.93} & \cellcolor{tabgray}{\bf 66.65$\pm$0.22} & \cellcolor{tabgray}{\bf 78.44$\pm$0.26} & \cellcolor{tabgray}{\bf 75.37$\pm$0.14} & \cellcolor{tabgray}{\bf 74.33$\pm$0.23} & \cellcolor{tabgray}{\bf 59.59$\pm$0.41} & 81.52$\pm$0.21 \\
    \midrule

    \multicolumn{12}{c}{\textbf{Mahalanobis}}\\
    $k=1$ & & 91.04$\pm$2.12 & 62.39$\pm$4.22 & 85.12$\pm$1.90 & 69.27$\pm$2.03 & 83.89$\pm$1.77 & 71.57$\pm$1.30 & 53.45$\pm$8.17 & 67.15$\pm$6.61 \\
    $k=2$ & & 88.68$\pm$2.96 & \cellcolor{tabgray}{\bf 65.43$\pm$2.15} & \cellcolor{tabgray}{\bf83.44$\pm$3.27} & 72.48$\pm$4.09 & \cellcolor{tabgray}{\bf83.41$\pm$2.78} & 73.45$\pm$2.57 & 44.60$\pm$8.88 & 75.07$\pm$5.99 \\
    $k=3$ & & 83.25$\pm$6.50 & 72.39$\pm$3.72 & 82.77$\pm$5.00 & 76.03$\pm$4.03 & 83.83$\pm$3.40 & 74.50$\pm$2.54 & \cellcolor{tabgray}{\bf37.22$\pm$7.31} & \cellcolor{tabgray}{\bf82.91$\pm$4.23} \\
    $k=4$ & & \cellcolor{tabgray}{\bf81.49$\pm$2.78} & 72.42$\pm$2.48 & 82.59$\pm$4.45 & \cellcolor{tabgray}{\bf76.15$\pm$3.11} & 84.14$\pm$3.55 & \cellcolor{tabgray}{\bf74.44$\pm$2.07} & 39.01$\pm$7.63 & 81.81$\pm$5.44 \\
    \midrule
    
    \multicolumn{12}{c}{\textbf{RankFeat}}\\
    $k=1$ & & 63.06$\pm$13.98 & 84.15$\pm$3.34 & 54.87$\pm$17.18 & 86.58$\pm$4.57 & 65.86$\pm$12.20 & 80.71$\pm$3.99 & 37.87$\pm$10.59 & 91.86$\pm$1.51 \\
    $k=2$ & & 49.18$\pm$13.04 & 89.67$\pm$2.61 & 45.30$\pm$11.30 & 89.80$\pm$2.57 & 59.64$\pm$8.77 & 83.70$\pm$2.61 & 23.21$\pm$2.71 & \cellcolor{tabgray}{\bf 95.04$\pm$0.10} \\
    $k=3$ & & 45.78$\pm$10.18 & 90.96$\pm$1.93 & 43.45$\pm$3.39 & 90.57$\pm$0.82 & \cellcolor{tabgray}{\bf 58.07$\pm$2.18} & 84.49$\pm$0.69 & \cellcolor{tabgray}{\bf 18.98$\pm$0.51} & 95.98$\pm$0.25 \\
    $k=4$ & & \cellcolor{tabgray}{\bf 40.92$\pm$5.41} & \cellcolor{tabgray}{\bf 91.82$\pm$0.99} & \cellcolor{tabgray}{\bf 38.68$\pm$3.01} & \cellcolor{tabgray}{\bf 91.51$\pm$0.52} & 54.84$\pm$2.31 & \cellcolor{tabgray}{\bf 85.38$\pm$0.49} & 16.44$\pm$0.77 & 96.43$\pm$0.36 \\
    \bottomrule
    \end{tabular}}
    \caption{The detection performance of ensembling $k$ independent modes (DenseNet121 trained on ImageNet-1K) on each OoD data set \textit{w.r.t} different types of OoD detectors. 
    Results with the {\bf lowest variances} are highlighted with bold fonts.}
    \label{tab:exp-ensemble-imgnet}
\end{table*}

As shown in Tab.\ref{tab:exp-ensemble-imgnet}, regarding the aforementioned 2 phenomena in the preceding subsection, we can find:
\begin{itemize}
    \item For the same OoD detector, mode ensemble could significantly reduce the detection variances and substantially improve the detection performance.
    For example, the mean FPR results of RankFeat \cite{song2022rankfeat} on iNaturalist \cite{van2018inaturalist} get improved from 63.06\% to 40.92\% with a variance reduction from 13.98 to 5.41.
    \item Such rather stable detection results brought by mode ensemble also benefit the evaluation of those OoD detectors and get rid of the uncertainties of single modes.
\end{itemize}

In addition, one can find disparities in performance enhancement brought by mode ensemble on different OoD detectors. 
RankFeat and Mahalanobis \cite{lee2018simple} exhibit considerable improvements, while MSP \cite{hendrycks2016baseline}, ODIN \cite{liang2018enhancing} and Energy \cite{liu2020energy} only show modest gains.
The reasons behind lie in the different types of these OoD detectors: features-based (RankFeat and Mahalanobis) and logits-based (MSP, ODIN and Energy).
The features of DNNs are generally viewed as latent representations of the inputs in a latent space, and thereby can contain richer information than the final logits of DNNs, as the latter is computed as the  probabilities distributed on the classification categories. In this way,
for multiple isolated modes of DNNs, the features learned by these modes can be more diverse and informative than the logits of these modes.
Accordingly, ensembling these features for Mahalanobis and RankFeat is naturally superior than ensembling these logits for MSP, ODIN and Energy, leading to the more substantial performance improvements

The advantages of mode ensemble, {\it i.e.}, variance reduction and performance improvements, are also widely reflected on the small-scale CIFAR10 benchmark for different types of OoD detectors.
Refer to Appendix \ref{app-sec:all-exp} for thorough results on mode ensemble for OoD detection.

\begin{table*}[ht]
    \centering
    \resizebox{\textwidth}{!}{
    \begin{tabular}{@{}c cc cc cc cc HH@{}}
    \toprule
    \multirow{3}{*}{Modes}  & \multicolumn{10}{c}{OoD data sets} \\
    & \multicolumn{2}{c}{iNaturalist} & \multicolumn{2}{c}{SUN} & \multicolumn{2}{c}{Places} & \multicolumn{2}{c}{Texture} & & \\
    & FPR$\downarrow$ & AUROC$\uparrow$ & FPR$\downarrow$ & AUROC$\uparrow$ & FPR$\downarrow$ & AUROC$\uparrow$ & FPR$\downarrow$ & AUROC$\uparrow$ & FPR$\downarrow$ & AUROC$\uparrow$ \\
    \midrule
    
    \multicolumn{11}{c}{\textbf{MSP}}\\
    mode-1 & \uwave{\bf 48.22} & \uwave{\bf 88.99} & \uwave{\bf 65.21} & 81.89 & \uwave{\bf 68.26} & \uwave{\bf 80.15} & 61.72 & 82.45 \\
    mode-2 & 47.41 & 89.27 & \underline{\bf 64.98} & \underline{\bf 82.13} & 68.69 & 80.39 & \underline{\bf 61.13} & \uwave{\bf 82.40} \\
    mode-3 & \underline{\bf 44.52} & \underline{\bf 89.78} & 65.12 & \uwave{\bf 81.82} & \underline{\bf 67.32} & \underline{\bf 80.75} & \uwave{\bf 61.88} & \underline{\bf 82.63} \\
    \cmidrule{2-11}
    AVG. single & 46.72$\pm$1.95 & 89.35$\pm$0.40 & 65.10$\pm$0.12 & 81.95$\pm$0.16 & 68.09$\pm$0.70 & 80.43$\pm$0.30 & 61.58$\pm$0.40 & 82.49$\pm$0.12 \\
    AVG. ensemble & \cellcolor{tabgray}{\bf42.70$\pm$0.92} & \cellcolor{tabgray}{\bf90.86$\pm$0.24} & \cellcolor{tabgray}{\bf64.38$\pm$0.39} & \cellcolor{tabgray}{\bf82.75$\pm$0.10} & \cellcolor{tabgray}{\bf67.29$\pm$0.26} & \cellcolor{tabgray}{\bf81.42$\pm$0.13} & \cellcolor{tabgray}{\bf57.54$\pm$0.39} & \cellcolor{tabgray}{\bf84.35$\pm$0.11} \\

    \midrule
    \multicolumn{11}{c}{\textbf{ODIN}}\\
    mode-1 & \uwave{\bf 45.30} & \uwave{\bf 87.41} & \uwave{\bf 64.34} & \uwave{\bf 77.42} & 68.26 & \uwave{\bf 73.78} & 56.58 & 81.29 \\
    mode-2 & 44.66 & 88.08 & 64.29 & \underline{\bf 78.09} & \uwave{\bf 69.36} & 74.68 & \uwave{\bf 56.74} & \uwave{\bf 81.25} \\
    mode-3 & \underline{\bf 41.28} & \underline{\bf 89.07} & \underline{\bf 63.77} & 77.80 & \underline{\bf 66.46} & \underline{\bf 75.38} & \underline{\bf 56.29} & \underline{\bf 81.90} \\
    \cmidrule{2-11}
    AVG. single & 43.75$\pm$2.16 & 88.19$\pm$0.84 & 64.13$\pm$0.32 & 77.77$\pm$0.34 & 68.03$\pm$1.46 & 74.61$\pm$0.80 & 56.54$\pm$0.23 & 81.48$\pm$0.36 \\
    AVG. ensemble & \cellcolor{tabgray}{\bf38.63$\pm$1.15} & \cellcolor{tabgray}{\bf91.00$\pm$0.35} & \cellcolor{tabgray}{\bf63.37$\pm$0.28} & \cellcolor{tabgray}{\bf79.77$\pm$0.17} & \cellcolor{tabgray}{\bf66.94$\pm$0.33} & \cellcolor{tabgray}{\bf76.93$\pm$0.36} & \cellcolor{tabgray}{\bf50.82$\pm$0.45} & \cellcolor{tabgray}{\bf85.01$\pm$0.19} \\

    \midrule
    \multicolumn{11}{c}{\textbf{Energy}}\\
    mode-1 & 57.93 & 81.43 & 73.11 & 69.69 & 76.81 & 64.47 & 58.58 & 78.54 \\
    mode-2 & \uwave{\bf 62.35} & \uwave{\bf 80.82} & \uwave{\bf 74.59} & \uwave{\bf 69.15} & \uwave{\bf 79.12} & \uwave{\bf 64.42} & \uwave{\bf 60.82} & \uwave{\bf 77.95} \\
    mode-3 & \underline{\bf 57.47} & \underline{\bf 83.55} & \underline{\bf 73.06} & \underline{\bf 70.37} & \underline{\bf 76.06} & \underline{\bf 66.72} & \underline{\bf 57.62} & \underline{\bf 79.39} \\
    \cmidrule{2-11}
    AVG. single & 59.25$\pm$2.69 & 81.93$\pm$1.43 & 73.59$\pm$0.87 & 69.74$\pm$0.61 & 77.33$\pm$1.59 & 65.20$\pm$1.31 & 59.01$\pm$1.64 & 78.63$\pm$0.72 \\
    AVG. ensemble & \cellcolor{tabgray}{\bf49.00$\pm$0.74} & \cellcolor{tabgray}{\bf88.20$\pm$0.43} & \cellcolor{tabgray}{\bf70.37$\pm$0.28} & \cellcolor{tabgray}{\bf74.26$\pm$0.22} & \cellcolor{tabgray}{\bf73.55$\pm$0.38} & \cellcolor{tabgray}{\bf69.95$\pm$0.55} & \cellcolor{tabgray}{\bf48.39$\pm$0.20} & \cellcolor{tabgray}{\bf84.76$\pm$0.26} \\


    \midrule
    \multicolumn{11}{c}{\textbf{Mahalanobis}}\\
    mode-1 & \underline{\bf94.92} & \underline{\bf54.64} & \underline{\bf96.43} & \underline{\bf45.48} & \underline{\bf97.14} & 43.48 & \uwave{\bf91.13} & \uwave{\bf60.91}\\
    mode-2 & \uwave{\bf99.64} & \uwave{\bf37.76} & \uwave{\bf98.23} & \uwave{\bf35.07} & \uwave{\bf98.84} & \uwave{\bf34.48} & \underline{\bf84.06} & \underline{\bf62.94}\\
    mode-3 & 98.15 & 39.87 & 97.48 & 45.17 & 97.78 & \underline{\bf44.37} & 87.15 & 61.77\\
    \cmidrule{2-11}
    AVG. single & 97.57$\pm$2.41 & 44.09$\pm$9.20 & 97.38$\pm$0.90 & 41.91$\pm$5.92 & 97.92$\pm$0.86 & 40.78$\pm$5.47 & 87.45$\pm$3.54 & 61.87$\pm$1.02 \\
    AVG. ensemble & \cellcolor{tabgray}{\bf94.52$\pm$4.31} & \cellcolor{tabgray}{\bf52.82$\pm$10.02} & \cellcolor{tabgray}{\bf95.70$\pm$1.85} & \cellcolor{tabgray}{\bf49.63$\pm$6.81} & \cellcolor{tabgray}{\bf96.71$\pm$1.10} & \cellcolor{tabgray}{\bf47.61$\pm$6.62} & \cellcolor{tabgray}{\bf84.52$\pm$9.02} & \cellcolor{tabgray}{\bf64.46$\pm$10.17} \\


    \midrule
    \multicolumn{11}{c}{\textbf{RankFeat}}\\
    mode-1 & 63.71 & 85.64 & 74.30 & 75.72 & 74.05 & 74.66 & 67.48 & \uwave{\bf 75.24} \\
    mode-2 & \uwave{\bf 69.54} & \uwave{\bf 80.87} & \uwave{\bf 78.46} & \uwave{\bf 75.06} & \uwave{\bf 83.74} & \uwave{\bf 70.18} & \underline{\bf 52.41} & \underline{\bf 85.63} \\
    mode-3 & \underline{\bf 61.25} & \underline{\bf 86.64} & \underline{\bf 71.18} & \underline{\bf 78.52} & \underline{\bf 71.55} & \underline{\bf 77.38} & \uwave{\bf 68.05} & 75.98 \\
    \cmidrule{2-11}
    AVG. single & 64.83$\pm$4.26 & 84.38$\pm$3.08 & 74.65$\pm$3.65 & 76.43$\pm$1.84 & 76.45$\pm$6.44 & 74.07$\pm$3.64 & 62.65$\pm$8.87 & 78.95$\pm$5.80 \\
    AVG. ensemble & \cellcolor{tabgray}{\bf54.06$\pm$5.83} & \cellcolor{tabgray}{\bf89.33$\pm$1.03} & \cellcolor{tabgray}{\bf70.91$\pm$2.56} & \cellcolor{tabgray}{\bf80.19$\pm$1.58} & \cellcolor{tabgray}{\bf72.19$\pm$1.20} & \cellcolor{tabgray}{\bf78.17$\pm$1.01} & \cellcolor{tabgray}{\bf55.77$\pm$7.63} & \cellcolor{tabgray}{\bf83.15$\pm$4.02} \\
    
    \bottomrule
    \end{tabular}}
    \caption{The detection performance of each of the 3 modes of T2T-ViT-14 trained from scratch on ImageNet-1K \textit{w.r.t} 5 OoD detectors: MSP, ODIN, Energy, Mahalanobis and RankFeat. The \underline{\bf best} and \uwave{\bf worst} results of each mode are highlighted.}
    \label{tab:exp-ensemble-imgnet-t2tvit14}
\end{table*}

\paragraph{Results on ViT models} In recent years, the Vision Transformer (ViT, \cite{dosovitskiy2020image}) has been a powerful network structure in various tasks \cite{peebles2023scalable}.
Therefore, experiments on ViT models are executed to further verify our observations on the performance of isolated modes in OoD detection.
Table \ref{tab:exp-ensemble-imgnet-t2tvit14} illustrates the detection performance of both single modes and mode ensemble ($k=2$), where each single mode is of T2T-ViT-14 \cite{yuan2021tokens} and trained on ImageNet-1K \cite{deng2009imagenet} from scratch.

As shown in Tab.\ref{tab:exp-ensemble-imgnet-t2tvit14}, due to the powerful generalization ability of T2T-ViT-14, there exist even smaller detection variances of multiple isolated modes, especially for those logits-based detectors (MSP \cite{hendrycks2016baseline}, ODIN \cite{liang2018enhancing} and Energy \cite{liu2020energy}).
Even though, those features-based detectors, especially RankFeat \cite{song2022rankfeat}, still hold very high variances among independent modes.
On the other hand, mode ensemble always brings variance reduction and substantial performance improvements for different OoD detectors.
In short, results in Tab.\ref{tab:exp-ensemble-imgnet-t2tvit14} again validate the high detection variances of single modes and the substantial improvements of mode ensemble in the case of ViT models.

\subsection{In-depth empirical discussions on mode ensemble}
\label{sec:exp-in-depth}

\subsubsection{On the independence of ensembling modes}
Ablation studies are provided to show that the ensemble of those modes, which are from the same trajectory and thus are not independent, can not bring OoD detection performance improvements.
\label{sec:exp-independ}

\begin{figure}[ht]
    \centering
    \includegraphics[width=0.7\linewidth]{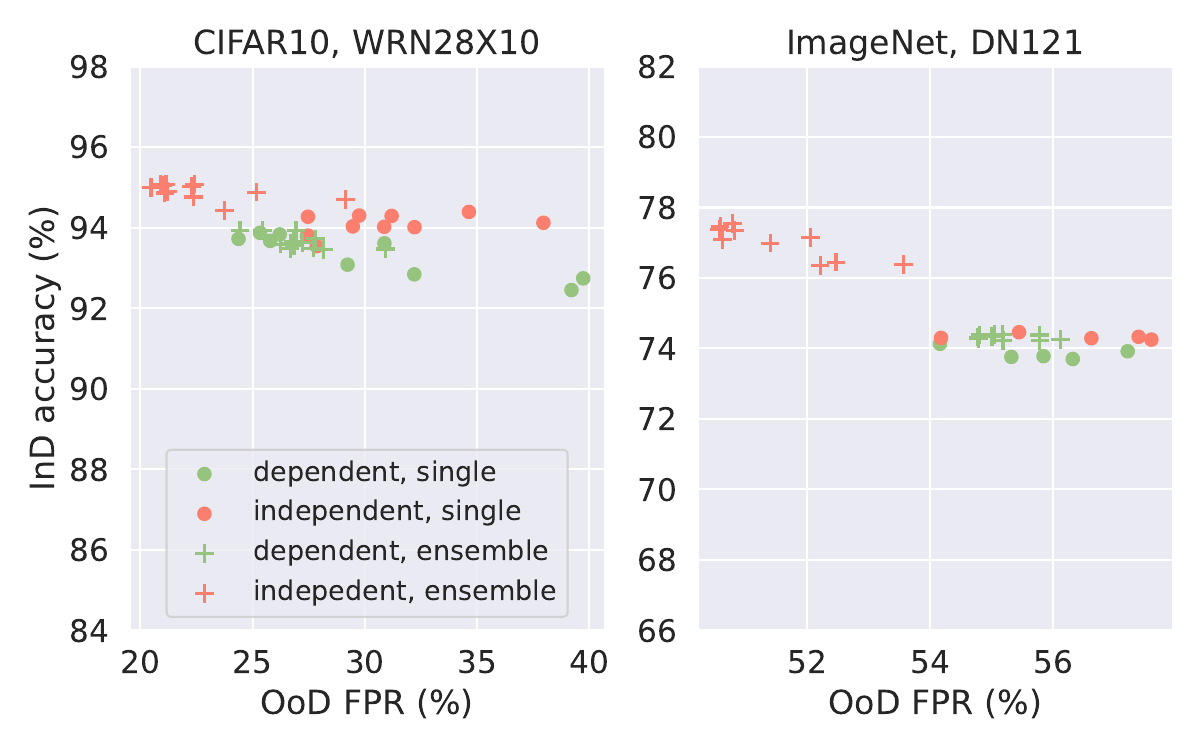}
    \caption{Ablation studies on the independence of modes. Each data point indicates the results of one single mode (circle dots) or ensembled modes (plus markers) with InD accuracy (y-axis) and average detection FPR over multiple OoD data sets (x-axis) achieved by Energy \cite{liu2020energy}.}
    \label{fig:exp-ablation}
\end{figure}

\paragraph{Setups}
Given a well-trained mode $f_{s_n}$ with parameters $\boldsymbol{\omega}_{s_n}\in\mathbb{R}^K$ \textit{w.r.t} a random seed $s_n$, we obtain $M$ modes $f_{s_n}^{(i)},i=1,\cdots,M$ by sampling from the subspace of $f_{s_n}$ as follows:
\begin{equation}
\label{eq:ablation}
\boldsymbol{\omega}_{s_n}^{(i)}=\boldsymbol{\omega}_{s_n}+r\cdot\frac{\boldsymbol{u}}{\|\boldsymbol{u}\|},i=1,\cdots,M,
\end{equation}
where $\boldsymbol{u}\in\mathcal{N}(0^K,1^K)$ and $r$ denotes a random step size.
In this way, the obtained $M$ modes $f_{s_n}^{(i)}$ are not independent with each other and are concentrated around $f_{s_n}$.
Comparison results between ensembling such modes and ensembling the independent modes on InD recognition accuracy and OoD detection FPR values are shown in Fig.\ref{fig:exp-ablation}.

\paragraph{Results}
As shown in Fig.\ref{fig:exp-ablation}, on the one hand, mode ensemble could bring higher InD accuracy and lower OoD FPR values for both the 2 types of modes that are isolated with each other or are all sampled from the same subspace.
On the other hand, ensembling those independent modes clearly shows better generalization performance on InD data and stronger detection performance on OoD data than ensembling those modes that are not isolated.
From the perspective of loss landscape, those isolated modes {\it w.r.t} different trajectories hold much more diversified information than those modes concentrated around the same trajectory, thereby leading to better ensemble performance in both InD generalization and OoD detection.

\subsubsection{On the model capacity of ensembling modes}
Aside from the enhanced detection performance, ensembling multiple modes simultaneously increases model capacities.
Therefore, it is meaningful to compare the performance of the mode ensemble with single models having an equivalent number of parameters, in order to provide insights into whether increasing model complexity is a viable alternative to enhancing OoD detection.
Accordingly, we design the following experiments on CIFAR10 and ImageNet-1K benchmarks.

\paragraph{Setups}
As shown in Tab.\ref{tab:model-capacity} below, when taking CIFAR10 \cite{krizhevsky2009learning} as the InD data, the number of parameters of the three ResNet18 (RN18, \cite{he2016deep}) networks is nearly equivalent with that of one Wide ResNet28X10 (WRN28X10, \cite{zagoruyko2016wide}) network.
When taking ImageNet-1K \cite{deng2009imagenet} as the InD data, the number of parameters of three DenseNet121 (DN121, \cite{huang2017densely}) networks is nearly equivalent with that of one ResNet50 (RN50, \cite{he2016deep}) network.
Therefore, in the experiments, the detection performance of ensembling $k=3$ modes of RN18 and DN121 is compared with that of a single WRN28X10 and RN50, respectively, whose detailed results are given in Tab.\ref{tab:exp-model-capacity-imgnet} and Tab.\ref{tab:exp-model-capacity-c10}.

\begin{table}[ht]
    \centering
    \scalebox{0.88}{
    \begin{tabular}{c|c|cc}
    \toprule
    data set & network & \# of modes & total capacity \\
    \midrule
    \multirow{2}{*}{CIFAR10} & RN18 & $k=3$ & 11,173,962 {\bf $\times3$} \\
    & WRN28X10 & $k=1$ & 36,489,290 \\
    \midrule
    \multirow{2}{*}{ImageNet-1K} & RN50 & $k=1$ & 25,557,032 \\
    & DN121 & $k=3$ & 7,978,856 {\bf $\times3$}\\
    \bottomrule
    \end{tabular}}
    \caption{Model capacities of different network structures.}
    \label{tab:model-capacity}
\end{table}

\paragraph{Results}
Taking the detection results on the large-scale ImageNet-1K benchmark in Tab.\ref{tab:exp-model-capacity-imgnet} as an example, despite the similar numbers of parameters, increasing model complexity does not necessarily guarantee the enhancement of OoD detection performance, which varies across different data sets and OoD detectors.
\begin{itemize}
    \item For the GradNorm detector \cite{huang2021importance}, a single mode of RN50 generally outperforms the ensemble of three modes of DN121 in detection performance under all the OoD data sets.
    \item Meanwhile, for the other detectors, especially the RankFeat \cite{song2022rankfeat}, the ensemble of three DN121 exhibits much better OoD detection performance than that of the single RN50.
\end{itemize}
The detection results on the small-scale CIFAR10 benchmark shown in Tab.\ref{tab:exp-model-capacity-c10} also present similar phenomena.
In short, simply increasing the model capacities cannot necessarily bring OoD detection performance improvements, as the detection performance is affected by multiple factors, such as the network structures, the tested OoD data, and the chosen detectors and ensemble strategies.

\begin{table*}[ht]
    \centering
    \resizebox{\textwidth}{!}{
    \begin{tabular}{c H cc cc cc cc HH}
    \toprule
    \multirow{3}{*}{\makecell[c]{Ensemble\\of $k$ modes}} & \multirow{3}{*}{InD ACC.}  & \multicolumn{10}{c}{OoD data sets} \\
    & & \multicolumn{2}{c}{iNaturalist} & \multicolumn{2}{c}{SUN} & \multicolumn{2}{c}{Places} & \multicolumn{2}{c}{Textures} & & \\
    & & FPR$\downarrow$ & AUROC$\uparrow$ & FPR$\downarrow$ & AUROC$\uparrow$ &
    FPR$\downarrow$ & AUROC$\uparrow$ & FPR$\downarrow$ & AUROC$\uparrow$ & FPR$\downarrow$ & AUROC$\uparrow$ \\
    \midrule

    \multicolumn{12}{c}{\textbf{MSP}}\\
    $k=1$ RN50 & & 56.86$\pm$1.76 & 87.50$\pm$0.45 & 70.75$\pm$0.55 & 80.92$\pm$0.30 & 72.65$\pm$0.25 & 80.32$\pm$0.12 & 68.01$\pm$1.48 & 80.45$\pm$0.61 \\
    $k=3$ DN121 & & 53.63$\pm$0.79 & 88.76$\pm$0.16 & 68.58$\pm$0.45 & 82.18$\pm$0.14 & 70.63$\pm$0.33 & 81.35$\pm$0.20 & 63.53$\pm$1.01 & 81.97$\pm$0.21 \\

    \midrule
    
    \multicolumn{12}{c}{\textbf{ODIN}}\\
    $k=1$ RN50 & & 54.09$\pm$1.52 & 90.04$\pm$0.32 & 59.67$\pm$0.75 & 86.47$\pm$0.44 & 63.73$\pm$0.79 & 84.74$\pm$0.28 & 55.95$\pm$1.69 & 85.81$\pm$0.64 \\
    $k=3$ DN121 & & 48.15$\pm$1.95 & 91.84$\pm$0.34 & 55.86$\pm$1.16 & 87.64$\pm$0.20 & 60.90$\pm$0.82 & 85.71$\pm$0.26 & 47.88$\pm$0.26 & 88.89$\pm$0.11 \\
    \midrule
    
    \multicolumn{12}{c}{\textbf{Energy}}\\
    $k=1$ RN50 & & 57.14$\pm$1.45 & 89.38$\pm$0.32 & 57.78$\pm$1.05 & 86.75$\pm$0.47 & 62.56$\pm$0.64 & 84.83$\pm$0.31 & 54.18$\pm$1.72 & 86.03$\pm$0.66 \\
    $k=3$ DN121 & & 49.65$\pm$1.88 & 91.43$\pm$0.41 & 52.68$\pm$1.03 & 88.03$\pm$0.22 & 58.96$\pm$0.69 & 85.87$\pm$0.28 & 43.95$\pm$0.52 & 89.53$\pm$0.11 \\
    \midrule
    
    \multicolumn{12}{c}{\textbf{GradNorm}}\\
    $k=1$ RN50 & & 56.87$\pm$4.22 & 83.87$\pm$1.71 & 64.23$\pm$2.37 & 79.38$\pm$1.15 & 72.42$\pm$1.49 & 75.28$\pm$0.86 & 60.02$\pm$2.96 & 79.25$\pm$1.23 \\
    $k=3$ DN121 & & 67.11$\pm$2.43 & 82.20$\pm$1.35 & 67.24$\pm$1.86 & 78.15$\pm$0.99 & 75.91$\pm$1.78 & 73.96$\pm$1.04 & 60.40$\pm$1.28 & 81.09$\pm$0.57 \\
    \midrule
    
    \multicolumn{12}{c}{\textbf{Mahalanobis}}\\
    $k=1$ RN50 & & 96.74$\pm$1.01 & 51.44$\pm$5.63 & 96.84$\pm$0.39 & 50.57$\pm$2.00 & 96.29$\pm$0.37 & 51.84$\pm$1.10 & 43.02$\pm$7.89 & 78.30$\pm$8.48 \\
    $k=3$ DN121 & & 83.08$\pm$3.89 & 71.71$\pm$5.53 & 82.28$\pm$4.06 & 75.31$\pm$2.27 & 83.55$\pm$3.54 & 74.49$\pm$1.74 & 41.35$\pm$4.06 & 79.32$\pm$2.80 \\
    \midrule
    
    \multicolumn{12}{c}{\textbf{RankFeat}}\\
    $k=1$ RN50 & & 87.91$\pm$8.24 & 63.17$\pm$11.47 & 82.25$\pm$6.77 & 73.38$\pm$7.74 & 85.79$\pm$5.13 & 68.94$\pm$6.86 & 63.83$\pm$9.05 & 79.96$\pm$5.41 \\
    $k=3$ DN121 & & 44.81$\pm$7.92 & 91.20$\pm$1.76 & 43.12$\pm$6.57 & 90.69$\pm$1.54 & 57.94$\pm$5.27 & 84.66$\pm$1.61 & 18.95$\pm$1.85 & 96.01$\pm$0.52 \\
    \bottomrule
    \end{tabular}}
    \caption{Comparisons on the detection performance between single modes and mode ensemble with nearly equivalent numbers of parameters on the {\bf ImageNet-1K} benchmark.}
    \label{tab:exp-model-capacity-imgnet}
\end{table*}

\begin{table*}[ht]
    \centering
    \resizebox{\textwidth}{!}{
    \begin{tabular}{c H cc cc cc cc cc HH}
    \toprule
    \multirow{3}{*}{\makecell[c]{Ensemble\\of $k$ modes}} & \multirow{3}{*}{InD ACC.}  & \multicolumn{12}{c}{OoD data sets} \\
    & & \multicolumn{2}{c}{SVHN} & \multicolumn{2}{c}{LSUN} & \multicolumn{2}{c}{iSUN} & \multicolumn{2}{c}{Texture} & \multicolumn{2}{c}{Places365} & & \\
    & & FPR$\downarrow$ & AUROC$\uparrow$ & FPR$\downarrow$ & AUROC$\uparrow$ &
    FPR$\downarrow$ & AUROC$\uparrow$ & FPR$\downarrow$ & AUROC$\uparrow$ & FPR$\downarrow$ & AUROC$\uparrow$ & FPR$\downarrow$ & AUROC$\uparrow$ \\
    \midrule

    \multicolumn{14}{c}{\textbf{MSP}}\\
    $k=1$ WRN28X10 & & 44.79$\pm$9.78 & 93.87$\pm$1.37 & 26.32$\pm$4.66 & 96.32$\pm$0.48 & 49.01$\pm$7.75 & 93.11$\pm$1.32 & 58.35$\pm$5.00 & 89.75$\pm$1.35 & 59.82$\pm$2.04 & 88.93$\pm$0.62 \\
    $k=3$ RN18 & & 67.09$\pm$3.11 & 88.48$\pm$1.74 & 46.83$\pm$3.00 & 93.28$\pm$0.58 & 55.09$\pm$1.26 & 92.02$\pm$0.60 & 58.36$\pm$1.13 & 90.59$\pm$0.31 & 59.05$\pm$1.38 & 89.32$\pm$0.10 \\

    \midrule
    
    \multicolumn{14}{c}{\textbf{ODIN}}\\
    $k=1$ WRN28X10 & & 25.83$\pm$10.93 & 95.66$\pm$1.82 & 5.52$\pm$1.25 & 98.91$\pm$0.20 & 24.38$\pm$5.94 & 95.98$\pm$0.78 & 47.00$\pm$6.76 & 89.81$\pm$2.12 & 44.65$\pm$4.41 & 90.45$\pm$1.16 \\
    $k=3$ RN18 & & 63.47$\pm$6.93 & 88.10$\pm$2.63 & 20.70$\pm$2.43 & 96.50$\pm$0.49 & 28.57$\pm$5.71 & 95.54$\pm$1.04 & 47.65$\pm$1.47 & 90.89$\pm$0.56 & 43.30$\pm$0.68 & 90.81$\pm$0.16 \\

    \midrule
    
    \multicolumn{14}{c}{\textbf{Energy}}\\
    $k=1$ WRN28X10 & & 25.36$\pm$11.32 & 95.76$\pm$1.80 & 5.97$\pm$1.23 & 98.84$\pm$0.21 & 29.65$\pm$7.06 & 95.32$\pm$1.03 & 48.56$\pm$6.76 & 89.61$\pm$2.05 & 44.89$\pm$4.51 & 90.50$\pm$1.14 \\
    $k=3$ RN18 & & 65.98$\pm$7.65 & 87.95$\pm$2.66 & 20.14$\pm$2.82 & 96.59$\pm$0.52 & 29.59$\pm$5.64 & 95.31$\pm$1.12 & 48.16$\pm$1.91 & 90.99$\pm$0.59 & 42.14$\pm$0.23 & 91.05$\pm$0.13 \\

    \midrule
    
    \multicolumn{14}{c}{\textbf{Mahalanobis}}\\
    $k=1$ WRN28X10 & & 12.07$\pm$18.53 & 97.30$\pm$3.94 & 73.88$\pm$7.96 & 76.89$\pm$4.60 & 6.77$\pm$1.87 & 98.26$\pm$0.60 & 24.54$\pm$8.47 & 92.46$\pm$3.18 & 85.08$\pm$2.29 & 69.46$\pm$2.21 \\
    $k=3$ RN18 & & 9.02$\pm$5.92 & 97.37$\pm$2.16 & 74.67$\pm$4.59 & 75.73$\pm$1.54 & 27.88$\pm$14.34 & 88.04$\pm$8.41 & 36.54$\pm$19.94 & 84.24$\pm$14.08 & 86.43$\pm$1.56 & 71.38$\pm$3.43 \\
    
    \midrule
    
    \multicolumn{14}{c}{\textbf{KNN}}\\
    $k=1$ WRN28X10 & & 23.35$\pm$4.76 & 96.46$\pm$0.63 & 20.23$\pm$2.34 & 96.74$\pm$0.36 & 26.87$\pm$7.25 & 95.62$\pm$1.18 & 36.07$\pm$5.58 & 94.15$\pm$0.98 & 46.43$\pm$3.07 & 90.44$\pm$0.76 \\
    $k=3$ RN18 & & 47.12$\pm$2.53 & 93.12$\pm$0.75 & 51.46$\pm$2.88 & 92.19$\pm$0.59 & 29.92$\pm$2.14 & 95.46$\pm$0.43 & 36.35$\pm$0.58 & 94.15$\pm$0.30 & 53.74$\pm$0.89 & 89.38$\pm$0.14 \\
    \bottomrule
    \end{tabular}}
    \caption{Comparisons on the detection performance between single modes and mode ensemble with nearly equivalent numbers of parameters on the {\bf CIFAR10} benchmark.}
    \label{tab:exp-model-capacity-c10}
\end{table*}

\subsubsection{On the sensitivity of the number of ensembling modes}

DNNs with different 
structures  have distinctive loss landscapes on OoD data, which can affect the detection performance when ensembling isolated modes in such loss landscapes. 
Thus, for different DNN structures, the chosen number of modes, {\it i.e.}, $k$, can  lead to varied performance boosts by the mode ensemble.
Accordingly, we further investigate the effects on $k$ under different DNN structures through the following experiments.

\paragraph{Setups} Varying numbers of ensembling modes $k$ are considered in experiments, and the corresponding detection performance brought by ensembling such modes is recorded under different DNN structures.
RN18 and WRN28X10 structures are taken for CIFAR10, while DN121 and T2T-ViT-14 structures are used for ImageNet-1K.
For the OoD detection in the experiments, we adopt the widely used and effective detection method Energy \cite{liu2020energy}.

\begin{figure*}[ht]
\centering
\includegraphics[width=1.0\linewidth]{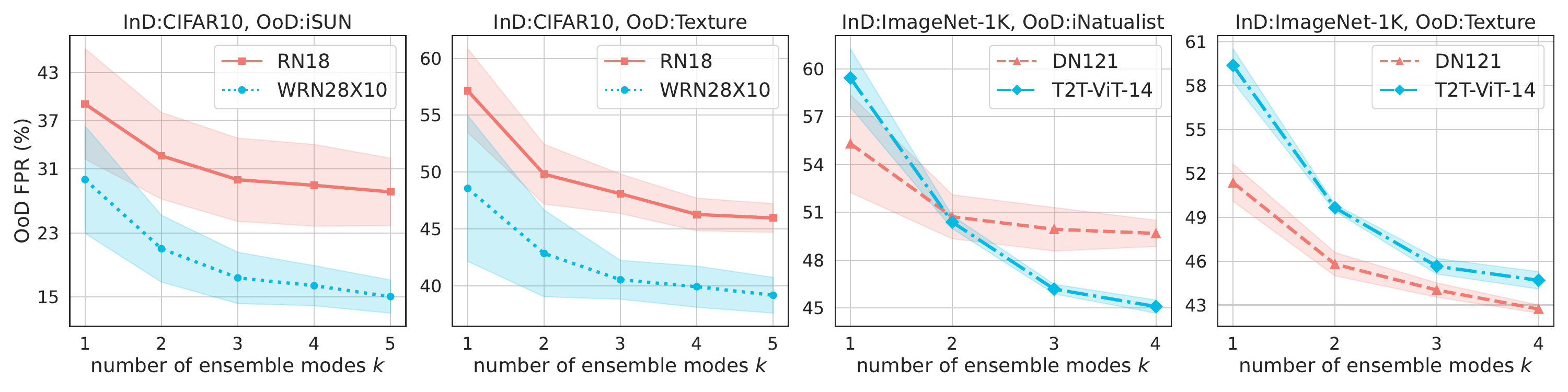}
\caption{
FPR of different DNN structures on CIFAR10 and ImageNet-1K benchmarks under varied numbers of ensembling modes $k$.
The shaded area indicates the standard deviations of FPR values of ensembling different modes.
}
\label{fig:sensitivity-K}
\end{figure*}

\paragraph{Results}
Figure \ref{fig:sensitivity-K} shows the mean and standard deviations of the FPR results from different DNN structures on CIFAR10 and ImageNet-1K benchmarks, as the number of ensembling modes $k$ increases.
\begin{itemize}
    \item  On the CIFAR10 benchmark, RN18 and WRN28X10 do not exhibit significant differences {\it w.r.t} the FPR metric in relation to the number of ensembling modes $k$.\\
    {\it In this experiment, the detection performance trending among the tested different DNN structures shows to be similar along with the variation of $k$.}
    \item  On the ImageNet-1K benchmark, as $k$ increases, the FPR drop, {\it i.e.}, the performance boost, varies distinctively between DN121 and T2T-ViT-14. 
    In Fig.\ref{fig:sensitivity-K}, the FPR values of the ViT-based structure (T2T-ViT-14) decrease more sharply than that of the DenseNet-based structure (DN121), where T2T-ViT-14 requires fewer numbers of modes for ensemble to achieve competitive or even better detection performances than DN121. It is also noticed that as $k$ increases, the standard deviation (the shaded area) of FPR from ensembling modes of T2T-ViT-14 shows to be much smaller than that from ensembling modes of DN121, which indicates {\it (i)} the stronger robustness of ViT agianst OoD and {\it (ii)} the decreased uncertainties with larger $k$.\\
   {\it In this experiment, the detection performance trending between the ViT and the DN121 differs in the varied FPR drops and the standard deviations of FPR along with increasing $k$, where fewer $k$ of  ensembling ViT yields more performance boosts.}
\end{itemize}

\subsubsection{On ensembling modes of different model structures}

\begin{table*}[ht]
    \centering
    \resizebox{\textwidth}{!}{
    \begin{tabular}{c H cc cc cc cc cc HH}
    \toprule
    \multirow{3}{*}{\makecell[c]{Ensemble\\of $k$ modes}} & \multirow{3}{*}{InD ACC.}  & \multicolumn{12}{c}{OoD data sets} \\
    & & \multicolumn{2}{c}{SVHN} & \multicolumn{2}{c}{LSUN} & \multicolumn{2}{c}{iSUN} & \multicolumn{2}{c}{Texture} & \multicolumn{2}{c}{Places365} & & \\
    & & FPR$\downarrow$ & AUROC$\uparrow$ & FPR$\downarrow$ & AUROC$\uparrow$ &
    FPR$\downarrow$ & AUROC$\uparrow$ & FPR$\downarrow$ & AUROC$\uparrow$ & FPR$\downarrow$ & AUROC$\uparrow$ & FPR$\downarrow$ & AUROC$\uparrow$ \\
    \midrule
    
    $k=1$ RN18 & & 62.67 & 88.40 & 22.73 & 96.02 & 39.92 & 92.96 & 49.54 & 90.37 & 45.73 & 89.64 \\
    $k=1$ WRN28X10 & & 21.94 & 96.38 & 3.85 & 99.21 & 25.64 & 95.86 & 8.51 & 92.12 & 47.43 & 90.18 \\
    \cmidrule{2-14}
    ENSEMBLE & & 27.49 & 95.60 & 6.09 & 98.76 & 19.66 & 96.81 & 36.05 & 93.54 & 40.28 & 91.77 \\

    
    \bottomrule
    \end{tabular}
    }
    \caption{Experiments on ensembling two modes of different structures (RN18 and WRN28X10) under the Energy detector on the CIFAR10 benchmark.}
    \label{tab:exp-ens-diff-arch-c10}
\end{table*}

\begin{table*}[ht]
    \centering
    \begin{tabular}{c H cc cc cc cc HH}
    \toprule
    \multirow{3}{*}{\makecell[c]{Ensemble\\of $k$ modes}} & \multirow{3}{*}{InD ACC.}  & \multicolumn{10}{c}{OoD data sets} \\
    & & \multicolumn{2}{c}{iNaturalist} & \multicolumn{2}{c}{SUN} & \multicolumn{2}{c}{Places} & \multicolumn{2}{c}{Textures} & & \\
    & & FPR$\downarrow$ & AUROC$\uparrow$ & FPR$\downarrow$ & AUROC$\uparrow$ &
    FPR$\downarrow$ & AUROC$\uparrow$ & FPR$\downarrow$ & AUROC$\uparrow$ & FPR$\downarrow$ & AUROC$\uparrow$ \\
    \midrule

    $k=1$ RN50 & & 56.06 & 89.64 & 56.88 & 87.37 & 61.85 & 85.21 & 55.53 & 85.63 \\
    $k=1$ DN121 & & 52.86 & 89.67 & 53.31 & 87.32 & 59.77 & 85.16 & 50.76 & 87.05 \\
    \cmidrule{2-12}
    ENSEMBLE & & 50.41 & 91.31 & 52.41 & 88.54 & 57.95 & 86.49 & 47.64 & 88.64 \\
    \bottomrule
    \end{tabular}
    \caption{Experiments on ensembling two modes of different structures (RN50 and DN121) under the Energy detector on the ImageNet-1K benchmark.}
    \label{tab:exp-ens-diff-arch-imgnet}
\end{table*}

In this section, a discussion on ensembling modes of different structures is provided.
When it comes to different types of OoD detectors, such as the {\it features}-based detector and the {\it logits}-based ones, there are several technical points to be taken into account for the ensemble.

\begin{itemize}
    \item For detectors relying on the {\it logits} of DNNs to determine the detection score, {\it e.g.}, the Energy \cite{liu2020energy}, it is convenient to directly ensemble the logits from different network structures, {\it e.g.}, a ResNet50 and a DenseNet121, since the outputs logits from the ResNet50 and the DenseNet121 hold the same dimension, and denote the probability distributions over the classification categories.
    \item For detectors relying on the {\it features} of DNNs to proceed the detection, {\it e.g.}, the KNN detector \cite{sun2022out}, 
    the ensemble requires the same dimensionality of the utilized features from different structures, {\it e.g.}, a ResNet50 and a DenseNet121.
    Besides, as features extracted by the ResNet50 and the DenseNet121 actually indicate distinctive  representations in their own latent spaces, ensembling such features might not guarantee detection performance improvements.
\end{itemize}

In the following Tab.\ref{tab:exp-ens-diff-arch-c10} and Tab.\ref{tab:exp-ens-diff-arch-imgnet}, we present two examples of ensembling two modes of different network structures under the Energy detector \cite{liu2020energy} (logits-based) on the CIFAR10 and ImageNet-1K benchmarks.
Detection performance improvements are also maintained by the mode ensemble on several OoD data sets.
Such mode ensemble for different structures can be easily extended to other detectors using logits to determine the detection score, such as MSP \cite{hendrycks2016baseline}, ODIN \cite{liang2018enhancing} and RankFeat \cite{song2022rankfeat}.
While for the detectors using features to calculate detection scores,  {\it e.g.}, the KNN detector \cite{sun2022out}, the penultimate features from the RN18 and WRN28X10 backbone have different dimensions, {\it i.e.}, 512 for RN18 and 640 for WRN28X10, the KNN detector cannot be proceeded in the mode ensemble, neither can another Mahalanobis detector \cite{lee2018simple} evaluated in our work.

\subsubsection{Comparisons with other ensemble detectors}
In the ablation studies in Sec.\ref{sec:exp-independ}, we have compared the detection performance between ensembling independent modes and dependent ones.
In this section, the proposed mode ensemble is further compared with another type of ensemble for OoD detection: ZODE \cite{xue2022boosting}.
Different from ensembling the output logits or features in DNNs of our method, ZODE proposes to perform ensemble on the p-values from multiple models, where the p-values are calculated from the detection scores.
Accordingly, the key difference between our mode ensemble and ZODE is,
\begin{itemize}
    \item In ours, the outputs (logits or features) from multiple modes are firstly get ensembled, then the detection score is computed based on the ensembled outputs to decide OoD predictions.
    \item In ZODE, the detection score of each individual model is firstly determined to calculate the p-value, then those p-values from multiple models get ensembled to decide OoD predictions.
\end{itemize}

Table \ref{tab:comparison-imgnet-zode} and Table \ref{tab:comparison-c10-zode} present the comparison results in OoD detection between ZODE \cite{xue2022boosting} and ours on ImageNet-1K and CIFAR10 benchmarks, respectively.
For fair comparisons, in each experiment, the $k$ modes for ensemble in our method are exactly the same models utilized in ZODE.
These $k$ models are randomly selected under each OoD detector, whose indices are also listed in the tables.

As shown in Tab.\ref{tab:comparison-imgnet-zode} and Tab.\ref{tab:comparison-c10-zode}, under different OoD detectors and different numbers of ensembled models, our mode ensemble method generally outperforms ZODE in terms of the average FPR values on multiple OoD data sets.
Such results indicate that ensembling the output features or logits from multiple DNNs might be a better choice than ensembling the scores from multiple DNNs in OoD detection, since features or logits could contain more useful and diversified information than the decision scores.

\begin{table*}[ht]
    \centering
    \resizebox{\textwidth}{!}{
    \begin{tabular}{c c cc cc cc cc cc}
    \toprule
    \multirow{3}{*}{\makecell[c]{Ensemble\\of $k$ modes}} & \multirow{3}{*}{Method}  & \multicolumn{10}{c}{OoD data sets} \\
    & & \multicolumn{2}{c}{iNaturalist} & \multicolumn{2}{c}{SUN} & \multicolumn{2}{c}{Places} & \multicolumn{2}{c}{Textures} & \multicolumn{2}{c}{\bf AVERAGE} \\
    & & FPR$\downarrow$ & AUROC$\uparrow$ & FPR$\downarrow$ & AUROC$\uparrow$ &
    FPR$\downarrow$ & AUROC$\uparrow$ & FPR$\downarrow$ & AUROC$\uparrow$ & FPR$\downarrow$ & AUROC$\uparrow$ \\
    \midrule

    \multicolumn{12}{c}{\textbf{MSP}}\\
    \multirow{2}{*}{\makecell[c]{$k=2$\\(mode-2,5)}} & ZODE & 56.40 & {\bf 88.68} & 69.63 & {\bf 82.28} & 71.70 & {\bf 81.43} & 63.58 & {\bf 82.75} & 65.33 & {\bf 83.79} \\
    & ours & \cellcolor{tabgray}{\bf55.85} & \cellcolor{tabgray}{88.01} & \cellcolor{tabgray}{\bf68.44} & \cellcolor{tabgray}{81.99} & \cellcolor{tabgray}{\bf70.78} & \cellcolor{tabgray}{81.21} & \cellcolor{tabgray}{\bf63.42} & \cellcolor{tabgray}{81.60} & \cellcolor{tabgray}{\bf64.62} & \cellcolor{tabgray}{83.20} \\
    \cmidrule{2-12}
    \multirow{2}{*}{\makecell[c]{$k=4$\\(mode-1,2,3,4)}} & ZODE & 56.02 & {\bf 90.00} & 70.80 & {\bf 82.81} & 73.45 & {\bf 81.89} & 64.13 & {\bf 83.96} & 66.10 & {\bf 84.67} \\
    & ours & \cellcolor{tabgray}{\bf52.86} & \cellcolor{tabgray}{89.06} & \cellcolor{tabgray}{\bf68.65} & \cellcolor{tabgray}{82.35} & \cellcolor{tabgray}{\bf70.81} & \cellcolor{tabgray}{81.54} & \cellcolor{tabgray}{\bf63.99} & \cellcolor{tabgray}{82.06} & \cellcolor{tabgray}{\bf64.08} & \cellcolor{tabgray}{83.75} \\

    \midrule
    \multicolumn{12}{c}{\textbf{ODIN}}\\
    \multirow{2}{*}{\makecell[c]{$k=2$\\(mode-1,5)}} & ZODE & 57.13 & 90.15 & 60.61 & 86.79 & 65.24 & 84.69 & 51.70 & 88.22 & 58.67 & 87.46 \\
    & ours & \cellcolor{tabgray}{\bf52.77} & \cellcolor{tabgray}{\bf90.58} & \cellcolor{tabgray}{\bf58.28} & \cellcolor{tabgray}{\bf86.92} & \cellcolor{tabgray}{\bf62.15} & \cellcolor{tabgray}{\bf84.94} & \cellcolor{tabgray}{\bf49.17} & \cellcolor{tabgray}{\bf88.37} & \cellcolor{tabgray}{\bf55.59} & \cellcolor{tabgray}{\bf87.70} \\
    \cmidrule{2-12}
    \multirow{2}{*}{\makecell[c]{$k=4$\\(mode-2,3,4,5)}} & ZODE & 56.38 & 91.65 & 60.86 & 87.82 & 65.70 & 85.93 & 51.45 & {\bf 89.06} & 58.60 & 88.62 \\
    & ours & \cellcolor{tabgray}{\bf48.04} & \cellcolor{tabgray}{\bf92.03} & \cellcolor{tabgray}{\bf55.26} & \cellcolor{tabgray}{\bf87.88} & \cellcolor{tabgray}{\bf60.09} & \cellcolor{tabgray}{\bf86.07} & \cellcolor{tabgray}{\bf46.93} & \cellcolor{tabgray}{89.05} & \cellcolor{tabgray}{\bf52.58} & \cellcolor{tabgray}{\bf88.76} \\

    \midrule
    \multicolumn{12}{c}{\textbf{Energy}}\\
    \multirow{2}{*}{\makecell[c]{$k=2$\\(mode-2,5)}} & ZODE & 57.55 & 89.78 & 55.38 & 87.72 & 61.16 & 85.67 & 47.62 & 88.65 & 55.43 & 87.96 \\
    & ours & \cellcolor{tabgray}{\bf53.22} & \cellcolor{tabgray}{\bf90.39} & \cellcolor{tabgray}{\bf52.64} & \cellcolor{tabgray}{\bf87.88} & \cellcolor{tabgray}{\bf58.32} & \cellcolor{tabgray}{\bf85.94} & \cellcolor{tabgray}{\bf44.66} & \cellcolor{tabgray}{\bf88.86} & \cellcolor{tabgray}{\bf52.21} & \cellcolor{tabgray}{\bf88.27} \\
    \cmidrule{2-12}
    \multirow{2}{*}{\makecell[c]{$k=4$\\(mode-1,2,3,4)}} & ZODE & 56.27 & 91.12 & 57.95 & 88.08 & 64.50 & 85.79 & 48.42 & 89.61 & 56.79 & 88.65 \\
    & ours & \cellcolor{tabgray}{\bf48.14} & \cellcolor{tabgray}{\bf91.84} & \cellcolor{tabgray}{\bf52.26} & \cellcolor{tabgray}{\bf88.25} & \cellcolor{tabgray}{\bf58.87} & \cellcolor{tabgray}{\bf86.08} & \cellcolor{tabgray}{\bf43.07} & \cellcolor{tabgray}{\bf89.85} & \cellcolor{tabgray}{\bf50.59} & \cellcolor{tabgray}{\bf89.01} \\

    \midrule
    \multicolumn{12}{c}{\textbf{Mahalanobis}}\\
    \multirow{2}{*}{\makecell[c]{$k=2$\\(mode-1,4)}} & ZODE & 92.00 & 61.54 & 88.50 & 68.85 & 87.45 & 71.05 & 44.40 & 74.42 & 78.09 & 68.97 \\
    & ours & \cellcolor{tabgray}{\bf87.65} & \cellcolor{tabgray}{\bf65.43} & \cellcolor{tabgray}{\bf81.10} & \cellcolor{tabgray}{\bf74.54} & \cellcolor{tabgray}{\bf81.20} & \cellcolor{tabgray}{\bf75.02} & \cellcolor{tabgray}{\bf39.43} & \cellcolor{tabgray}{\bf78.06} & \cellcolor{tabgray}{\bf72.35} & \cellcolor{tabgray}{\bf73.26} \\
    \cmidrule{2-12}
    \multirow{2}{*}{\makecell[c]{$k=4$\\(mode-1,3,4,5)}} & ZODE & 94.10 & 62.72 & 90.77 & 69.80 & 89.58 & 72.23 & 46.17 & 74.44 & 80.16 & 69.80 \\
    & ours & \cellcolor{tabgray}{\bf79.60} & \cellcolor{tabgray}{\bf71.33} & \cellcolor{tabgray}{\bf77.54} & \cellcolor{tabgray}{\bf79.12} & \cellcolor{tabgray}{\bf80.09} & \cellcolor{tabgray}{\bf76.72} & \cellcolor{tabgray}{\bf33.16} & \cellcolor{tabgray}{\bf85.21} & \cellcolor{tabgray}{\bf67.60} & \cellcolor{tabgray}{\bf78.10} \\

    \bottomrule
    \end{tabular}
    }
    \caption{Comparisons between ZODE and ours on ensembling $k$ modes ({\bf DenseNet121} trained on {\bf ImageNet-1K}) on each OoD data set \textit{w.r.t} different OoD detectors.}
    \label{tab:comparison-imgnet-zode}
\end{table*}

\begin{table*}[ht]
    \centering
    \resizebox{\textwidth}{!}{
    \begin{tabular}{c c cc cc cc cc cc cc}
    \toprule
    \multirow{3}{*}{\makecell[c]{Ensemble\\of $k$ modes}} & \multirow{3}{*}{Method}  & \multicolumn{12}{c}{OoD data sets} \\
    & & \multicolumn{2}{c}{SVHN} & \multicolumn{2}{c}{LSUN} & \multicolumn{2}{c}{iSUN} & \multicolumn{2}{c}{Texture} & \multicolumn{2}{c}{Places365} & \multicolumn{2}{c}{\bf AVERAGE} \\
    & & FPR$\downarrow$ & AUROC$\uparrow$ & FPR$\downarrow$ & AUROC$\uparrow$ &
    FPR$\downarrow$ & AUROC$\uparrow$ & FPR$\downarrow$ & AUROC$\uparrow$ & FPR$\downarrow$ & AUROC$\uparrow$ & FPR$\downarrow$ & AUROC$\uparrow$ \\
    \midrule

    \multicolumn{14}{c}{\textbf{MSP}}\\
    \multirow{2}{*}{\makecell[c]{$k=2$\\(mode-2,8)}} & ZODE & 74.47 & {\bf 86.91} & 50.68 & 92.90 & 52.44 & {\bf 93.08} & 62.57 & {\bf 90.06} & 62.22 & 88.89 & 60.48 & 90.37 \\
    & ours & \cellcolor{tabgray}{\bf72.64} & \cellcolor{tabgray}{86.88} & \cellcolor{tabgray}{\bf49.05} & \cellcolor{tabgray}{\bf93.12} & \cellcolor{tabgray}{\bf51.88} & \cellcolor{tabgray}{92.98} & \cellcolor{tabgray}{\bf61.15} & \cellcolor{tabgray}{90.01} & \cellcolor{tabgray}{\bf60.08} & \cellcolor{tabgray}{\bf89.09} & \cellcolor{tabgray}{\bf58.96} & \cellcolor{tabgray}{\bf90.42} \\
    \cmidrule{2-14}
    \multirow{2}{*}{\makecell[c]{$k=4$\\(mode-4,6,7,8)}} & ZODE & 69.85 & 89.49 & {\bf 44.71} & {\bf 93.67} & {\bf 49.76} & {\bf 93.28} & 58.24 & {\bf 91.23} & 58.77 & 89.50 & 56.27 & {\bf 91.43} \\
    & ours & \cellcolor{tabgray}{\bf64.69} & \cellcolor{tabgray}{\bf89.58} & \cellcolor{tabgray}{45.34} & \cellcolor{tabgray}{93.54} & \cellcolor{tabgray}{52.96} & \cellcolor{tabgray}{92.49} & \cellcolor{tabgray}{\bf57.36} & \cellcolor{tabgray}{91.05} & \cellcolor{tabgray}{\bf 58.21} & \cellcolor{tabgray}{\bf89.53} & \cellcolor{tabgray}{\bf55.71} & \cellcolor{tabgray}{91.24} \\

    \midrule
    \multicolumn{14}{c}{\textbf{ODIN}}\\
    \multirow{2}{*}{\makecell[c]{$k=2$\\(mode-3,7)}} & ZODE & 75.74 & 83.48 & {\bf 23.52} & {\bf 96.13} & {\bf 34.53} & {\bf 95.05} & 54.36 & 89.57 & 48.14 & {\bf 89.91} & 47.26 & {\bf 90.83} \\
    & ours & \cellcolor{tabgray}{\bf72.86} & \cellcolor{tabgray}{\bf83.66} & \cellcolor{tabgray}{24.70} & \cellcolor{tabgray}{95.75} & \cellcolor{tabgray}{36.90} & \cellcolor{tabgray}{94.65} & \cellcolor{tabgray}{\bf50.92} & \cellcolor{tabgray}{\bf89.93} & \cellcolor{tabgray}{\bf47.31} & \cellcolor{tabgray}{89.81} & \cellcolor{tabgray}{\bf46.64} & \cellcolor{tabgray}{90.76} \\
    \cmidrule{2-14}
    \multirow{2}{*}{\makecell[c]{$k=4$\\(mode-6,7,8,9)}} & ZODE & 75.50 & 88.98 & 16.60 & {\bf 97.57} & {\bf 21.89} & {\bf 97.01} & 51.93 & 91.63 & 45.72 & {\bf 91.36} & 42.33 & 93.31 \\
    & ours & \cellcolor{tabgray}{\bf63.13} & \cellcolor{tabgray}{\bf89.78} & \cellcolor{tabgray}{\bf15.36} & \cellcolor{tabgray}{97.30} & \cellcolor{tabgray}{23.08} & \cellcolor{tabgray}{96.42} & \cellcolor{tabgray}{\bf44.10} & \cellcolor{tabgray}{\bf91.87} & \cellcolor{tabgray}{\bf40.80} & \cellcolor{tabgray}{91.34} & \cellcolor{tabgray}{\bf37.29} & \cellcolor{tabgray}{\bf93.34} \\

    \midrule
    \multicolumn{14}{c}{\textbf{Energy}}\\
    \multirow{2}{*}{\makecell[c]{$k=2$\\(mode-4,8)}} & ZODE & 72.76 & {\bf 87.78} & 24.31 & {\bf 96.36} & 27.22 & 96.00 & 58.09 & 90.03 & 48.01 & 90.48 & 46.08 & {\bf 92.13} \\
    & ours & \cellcolor{tabgray}{\bf71.68} & \cellcolor{tabgray}{87.22} & \cellcolor{tabgray}{\bf21.85} & \cellcolor{tabgray}{96.31} & \cellcolor{tabgray}{\bf23.84} & \cellcolor{tabgray}{\bf96.03} & \cellcolor{tabgray}{\bf51.05} & \cellcolor{tabgray}{\bf90.43} & \cellcolor{tabgray}{\bf44.08} & \cellcolor{tabgray}{\bf90.60} & \cellcolor{tabgray}{\bf42.50} & \cellcolor{tabgray}{92.12} \\
    \cmidrule{2-14}
    \multirow{2}{*}{\makecell[c]{$k=4$\\(mode-3,4,6,7)}} & ZODE & 72.85 & 88.73 & 23.14 & {\bf 96.63} & 39.98 & 94.67 & 55.32 & 91.38 & 47.53 & 91.03 & 47.76 & 92.49 \\
    & ours & \cellcolor{tabgray}{\bf61.06} & \cellcolor{tabgray}{\bf89.30} & \cellcolor{tabgray}{\bf19.83} & \cellcolor{tabgray}{96.55} & \cellcolor{tabgray}{\bf33.65} & \cellcolor{tabgray}{\bf94.80} & \cellcolor{tabgray}{\bf44.91} & \cellcolor{tabgray}{\bf91.76} & \cellcolor{tabgray}{\bf41.44} & \cellcolor{tabgray}{\bf91.13} & \cellcolor{tabgray}{\bf40.18} & \cellcolor{tabgray}{\bf92.71} \\

    \midrule
    \multicolumn{14}{c}{\textbf{Mahalanobis}}\\
    \multirow{2}{*}{\makecell[c]{$k=2$\\(mode-1,7)}} & ZODE & 21.73 & 95.53 & 64.32 & 84.29 & {\bf 11.94} & {\bf 96.49} & {\bf 50.82} & {\bf 76.88} & {\bf 86.90} & 69.68 & 47.14 & {\bf 84.57} \\
    & ours & \cellcolor{tabgray}{\bf17.55} & \cellcolor{tabgray}{\bf95.59} & \cellcolor{tabgray}{\bf55.33} & \cellcolor{tabgray}{\bf86.17} & \cellcolor{tabgray}{14.10} & \cellcolor{tabgray}{95.36} & \cellcolor{tabgray}{59.33} & \cellcolor{tabgray}{69.93} & \cellcolor{tabgray}{87.60} & \cellcolor{tabgray}{\bf69.93} & \cellcolor{tabgray}{\bf46.78} & \cellcolor{tabgray}{83.40} \\
    \cmidrule{2-14}
    \multirow{2}{*}{\makecell[c]{$k=4$\\(mode-1,3,6,7)}} & ZODE & 14.17 & 97.36 & {\bf 69.28} & {\bf 82.88} & {\bf 6.26} & {\bf 98.73} & 29.26 & {\bf 93.17} & 88.44 & 70.70 & 41.46 & {\bf 88.57} \\
    & ours & \cellcolor{tabgray}{\bf4.65} & \cellcolor{tabgray}{\bf98.90} & \cellcolor{tabgray}{76.00} & \cellcolor{tabgray}{74.98} & \cellcolor{tabgray}{10.57} & \cellcolor{tabgray}{97.66} & \cellcolor{tabgray}{\bf28.17} & \cellcolor{tabgray}{91.00} & \cellcolor{tabgray}{\bf86.71} & \cellcolor{tabgray}{\bf73.34} & \cellcolor{tabgray}{\bf41.22} & \cellcolor{tabgray}{87.18} \\
    
    \bottomrule
    \end{tabular}}
    \caption{Comparisons between ZODE and ours on ensembling $k$ modes ({\bf ResNet18} trained on {\bf CIFAR10}) on each OoD data set \textit{w.r.t} different OoD detectors.}
    \label{tab:comparison-c10-zode}
\end{table*}

\section{Related work}
\label{sec:related-work}
\subsection{OoD detection}
The task of OoD detection focuses on the detection ability of DNNs on data from an out distribution $\mathcal{P}_{\rm out}$ differing from the training distribution \cite{nguyen2015deep}.
Plenty of researches have been developed from various aspects, see reviews in \cite{yang2024generalized,shen2021towards}.

Aside from the outlined 3 types of detectors, logits-based, features-based and gradients-based, in Sec.\ref{sec:intro}, which are mainly post-hoc and OoD-agnostic, there exist other detectors that work on the training process via additional data or model modifications.
For example, G-ODIN \cite{hsu2020generalized} designs a two-branch network structure to decompose the output probabilities and to alleviate the over-confidence on OoD data.
An auto-encoder is introduced for OoD detection in an unsupervised way with the Stiefel-Restricted Kernel Machine in \cite{tonin2021unsupervised}.
\cite{yang2023full} discusses 2 different distribution shifts for the out distribution $\mathcal{P}_{\rm out}$ and proposes semantics information as the detection metric.

Related theoretical works in OoD detection mainly study single modes.
A theoretical framework for OoD generalization is formalized in \cite{ye2021towards}.
The probably approximated correct learning theory for OoD detection is investigated in \cite{fang2022out}.
Our theoretical work discusses the mode ensemble for OoD detection in a simple binary classification problem on Gaussian distributions.

In this work, we discover the uncertainty in OoD evaluations on the loss landscape of isolated modes.
Interestingly, \cite{zhu2024rethinking} also discusses similar unreliability issues in OoD evaluations.
Different from our loss landscape perspective, the key viewpoint of \cite{zhu2024rethinking} is from an image-level view, and a human-centric OoD evaluation metric is proposed to alleviate the difficulty of detecting the distribution shift in input images.
Both our work and \cite{zhu2024rethinking} provide insights on the reliability issues of current OoD evaluations and offer corresponding viable solutions from different aspects.

\subsection{Mode, loss landscape and ensemble}
\paragraph{Mode and loss landscape} Analyzing the mode properties in the loss landscape of DNNs could date back to \cite{goodfellow2014qualitatively}, which shows that the loss along a straight line connecting the initialization and the final solution monotonically decreases.
Then, \cite{draxler2018essentially} and \cite{garipov2018loss} find low-loss curves that connect two independent modes in the loss landscape and \cite{wortsman2021learning} proposes a training algorithm to learn subspaces of DNNs in the forms of lines, Bezier curves and simplexes.
Mode connectivity has also inspired researches in other fields.
For example, \cite{zhao2020bridging,wang2023exploring} leverage mode connectivity for more robust DNNs against adversarial examples.
\cite{rame2022diverse} proposes weight-averaging on the independent modes for better OoD generalization.

\paragraph{Ensemble} Ensemble has been a useful technique to improve the model generalization performance on the InD test data in both classic machine learning \cite{zhou2012ensemble} and the deep learning \cite{lakshminarayanan2017simple}.
\cite{fort2019deep} visualizes and quantitatively measures the functional space diversity in the loss landscape to show how deep ensemble boosts the generalization performance on InD data.
Besides, the ensemble technique has been widely utilized in different fields, such as adversarial and certified robustness \cite{fang2024towards,horvath2021boosting,fang2022multi} and machinery fault diagnosis \cite{han2022out}.

\section{Conclusion and discussion}
\label{sec:conclusion}
In this work, starting from the loss landscape perspective, we observe that those independent modes, which are trained to achieve low-loss regions on InD data {\it w.r.t} different random seeds, yet yield significantly different loss landscapes of OoD data.
The research community has ignored the uncertainties brought by such diversities across isolated modes on the OoD loss landscape: (i) For the same OoD detector, independent modes could hold substantially varied detection performance, and (ii) For two OoD detectors, different modes could yield opposite performance.
Accordingly, to alleviate the high variances, we revisit the deep ensemble method to ensemble multiple independent modes so as to achieve stable results and substantially improve detecting OoD data, which also benefits the development and evaluation on OoD detectors in the sense of reduced variances.
Our experiments cover different types of OoD detectors, various network structures, small-scale and large-scale data sets with in-depth explorations on many aspects of the mode ensemble, which supports our viewpoint comprehensively and solidly together with the theoretical analysis on the reduced accuracy gap between InD and OoD data.

Nevertheless, the extra computation in training has been an inevitable limitation in ensemble methods, as sequentially training multiple DNNs is time-consuming.
One could mitigate this problem via more efficient training tricks, such as the parallel training for the base DNNs.
Besides, in the future work, more sophisticated ensemble strategies could be specifically devised for particular OoD detectors, on the basis of the unique properties of the detector and the provided baseline ensemble strategies of our work.

\section*{Acknowledgements}
This work is jointly supported by National Natural Science Foundation of China (62376155, 62376153), Shanghai Municipal Science and Technology Research Program (22511105600), and Shanghai Municipal Science and Technology Major Project (2021SHZDZX0102). 
Jie YANG and Xiaolin Huang are the corresponding authors.

\bibliographystyle{unsrt}  
\bibliography{main_reference}

\clearpage

\begin{appendices}
\section{Baseline OoD detectors and mode ensemble}
\label{app-sec:ood-baseline-ensemble}

We outline how the scoring function $S(\cdot)$ is designed in the selected baseline OoD detectors and elaborate the corresponding mode ensemble strategies over these detectors.
In the following, the outputs of the mode $f:\mathbb{R}^D\rightarrow\mathbb{R}^C$ are the logits of $C$-dimension, in correspondence to $C$ classes.

\textbf{MSP} \cite{hendrycks2016baseline} takes the maximum probability over the output logits as the scoring function.
Given a new sample $\boldsymbol{x}\in\mathbb{R}^D$, its MSP score \textit{w.r.t} a single mode $f(\cdot)$ is
\begin{equation}
\label{eq:msp}
S_{\rm MSP}(\boldsymbol{x})=\max\left({\rm softmax}(f(\boldsymbol{x}))\right).
\end{equation}

\textbf{Ensemble on MSP} adopts the average logits from $N$ modes $f_{s_i},i=1,\cdots,N$ to calcuate the maximum probability as the score for $\boldsymbol{x}$:
\begin{equation}
\label{eq:msp-ensemble}
S_{\rm MSP\text{-}ens}(\boldsymbol{x})=\max\left({\rm softmax}\left(\frac{1}{N}\sum_{i=1}^Nf_{s_i}(\boldsymbol{x})\right)\right).
\end{equation}

\textbf{ODIN} \cite{liang2018enhancing} introduces temperature scaling and adversarial examples into MSP and proposes the following score:
\begin{equation}
\label{eq:odin}
S_{\rm ODIN}(\boldsymbol{x})=\max\left({\rm softmax}(\frac{f(\boldsymbol{\bar x})}{T})\right),
\end{equation}
where $T$ denotes the temperature and $\boldsymbol{\bar x}$ denotes the perturbed adversarial example of $\boldsymbol{x}$.
Following the settings in \cite{liang2018enhancing,song2022rankfeat}, we set $T=1000$ and do not perturb $\boldsymbol{x}$ for the ImageNet-1K benchmark in experiments.

\textbf{Ensemble on ODIN} shares a similar scoring function with that of MSP-ensemble:
\begin{equation}
\label{eq:odin-ensemble}
S_{\rm ODIN\text{-}ens}(\boldsymbol{x})=\max\left({\rm softmax}(\frac{\sum_{i=1}^Nf_{s_i}(\boldsymbol{\bar x})}{N\cdot T})\right).
\end{equation}
The adversarial attack is executed individually on each mode $f_{s_i}$ and then the ODIN score is calculated on the average predictive logits on the perturbed inputs.

\textbf{Energy} \cite{liu2020energy} improves MSP via an energy function since energy is better aligned with the input probability density:
\begin{equation}
\label{eq:energy}
S_{\rm energy}(\boldsymbol{x})=\log\sum_{i=1}^C\exp(f^i(\boldsymbol{x})),
\end{equation}
where $f^i(\boldsymbol{x})$ denotes the $i$-th element in the $C$-dimension output logits.

\textbf{Ensemble on Energy} firstly averages the $N$ logits and then computes the energy score:
\begin{equation}
\label{eq:energy-ensemble}
\begin{aligned}
S_{\rm energy\text{-}ens}(\boldsymbol{x})=\log\sum_{i=1}^C\exp(f_{\rm ens}^i(\boldsymbol{x})),\quad f_{\rm ens}(\boldsymbol{x})=\frac{1}{N}\sum_{i=1}^Nf_{s_i}(\boldsymbol{x}).
\end{aligned}
\end{equation}

\textbf{Mahalanobis} \cite{lee2018simple} tries to model the network outputs at different layers as the mixture of multivariate Gaussian distributions and uses the Mahalanobis distance as the scoring function:
\begin{equation}
\label{eq:mahal}
S_{\rm mahal}(\boldsymbol{x})=\max_c\left(-(f(\boldsymbol{x})-\boldsymbol{\mu}_c)^\top\Sigma(f(\boldsymbol{x})-\boldsymbol{\mu}_c)\right),
\end{equation}
where $\boldsymbol{\mu}_c$ denotes the mean feature vector of class-$c$ and $\Sigma$ denotes the covariance matrix across classes.
In experiments, following the settings in \cite{lee2018simple,song2022rankfeat}, adversarial examples generated from 500 randomly-selected clean samples are involved to train the logistic regression, with a perturbation size 0.05 on CIFAR10 and 0.001 on ImageNet-1K.

\textbf{Ensemble on Mahalanobis} leverages the average output features  at the same layers in DNNs over $N$ modes and attacks the $N$ modes simultaneously to calculate the Mahalanobis score.
Details can be found in the released code.

\textbf{KNN} \cite{sun2022out} is a simple but time-consuming and memory-inefficient detector since it performs nearest neighbor search on the $\ell_2$-normalized penultimate features between the test sample and all the training samples.
The negative of the ($k$-th) shortest $\ell_2$ distance between the features $\boldsymbol{h}^*$ of a new sample $\boldsymbol{x}^*$ and all the training features $\boldsymbol{h}^i$ is set as the score:
\begin{equation}
\label{eq:KNN}
S_{\rm KNN}(\boldsymbol{x}^*)=-\min_{i:1,\cdots,n_{\rm tr}}\left\|\frac{\boldsymbol{h}^*}{\|\boldsymbol{h}^*\|_2}-\frac{\boldsymbol{h}^i}{\|\boldsymbol{h}^i\|_2}\right\|_2,
\end{equation}
where $\boldsymbol{h}$ denotes the penultimate features in the DNN, and $\boldsymbol{h}^i$ denotes the penultimate features in correspondence to the $i$-th training sample in the training set of size $n_{\rm tr}$.
The key of this detector is the $\ell_2$ normalization on the features.

\textbf{Ensemble on KNN} improves performance by replacing the penultimate features from one single mode with the average penultimate features from $N$ nodes:
\begin{equation}
\label{eq:KNN-ensemble}
\begin{aligned}
S_{\rm KNN\text{-}ens}(\boldsymbol{x}^*)
=-\min_{j:1,\cdots,n_{\rm tr}}
\left\|
\frac{\sum_{i=1}^N\boldsymbol{h}_{s_i}^*}{\|\sum_{i=1}^N\boldsymbol{h}_{s_i}^*\|_2}
-\frac{\sum_{i=1}^N\boldsymbol{h}_{s_i}^j}{\|\sum_{i=1}^N\boldsymbol{h}_{s_i}^j\|_2}
\right\|_2.
\end{aligned}
\end{equation}

\textbf{RankFeat} \cite{song2022rankfeat} removes the rank-1 matrix from each individual sample feature matrix $\mathbf{X}$ in the mini-batch during forward propagation in test, since the rank-1 feature drastically perturbs the predictions on OoD samples:
\begin{equation}
\begin{aligned}
\label{eq:rankfeat}
&\mathbf{X}=\mathbf{U}\mathbf{S}\mathbf{V}^\top,\\
&\mathbf{X}^\prime=\mathbf{X}-\boldsymbol{s}_1\boldsymbol{u}_1\boldsymbol{v}_1^\top.
\end{aligned}
\end{equation}
In Eq.\eqref{eq:rankfeat}, the singular value decomposition is firstly executed on the feature matrix $\mathbf{X}$ of an individual sample, leading to the left and right orthogonal singular vector matrices $\mathbf{U}$ and $\mathbf{V}$, and the rectangle diagonal singular value matrix $\mathbf{S}$.
Then, the rank-1 matrix is calculated based on the largest singular value $\boldsymbol{s}_1$ and the two corresponding singular vectors $\boldsymbol{u}_1$ and $\boldsymbol{v}_1$, and gets subtracted from the original $\mathbf{X}$.
Such removals are recommended at the 3rd and 4th blocks in DNNs.
Finally, an Energy score \cite{liu2020energy} is calculated on the resulting changed output logits.

\textbf{Ensemble on RankFeat} executes the rank-1 feature removing on each mode individually and then average the $N$ changed logits to compute the Energy score as Eq.\eqref{eq:energy-ensemble}.
Details can be found in the released code.

\textbf{GradNorm} \cite{huang2021importance} leverages the gradient information for OoD detection by calculating the $\ell_1$ norm of the gradients \textit{w.r.t} a KL divergence loss as the score of $\boldsymbol{x}$:
\begin{equation}
\label{eq:gradnorm}
S_{\rm GN}(\boldsymbol{x})=\left\|\frac{\partial\ {\rm KL}(\boldsymbol{u}\ \|\ {\rm softmax}(f(\boldsymbol{x})))}{\partial\ \boldsymbol{\omega}}\right\|_1,
\end{equation}
where $\boldsymbol{u}=[1/C,\cdots,1/C]\in\mathbb{R}^C$ and $\boldsymbol{\omega}$ is set as the weight parameters of the last full-connected layer in DNNs.

\textbf{Ensemble on GradNorm} firstly calculates the KL divergence between $\boldsymbol{u}$ and the softmax probability of the average logits of $N$ modes.
The final score for $\boldsymbol{x}$ is the average gradient norm over the selected parameters $\boldsymbol{\omega}_{s_i}$ in each mode:
\begin{equation}
\label{gradnorm-ensemble}
\begin{aligned}
S_{\rm GN\text{-}ens}(\boldsymbol{x})=\frac{1}{N}\sum_{i=1}^N\left\|\frac{\partial\ {\rm KL}(\boldsymbol{u}\ \|\ {\rm softmax}(f_{\rm ens}(\boldsymbol{x})))}{\partial\ \boldsymbol{\omega}_{s_i}}\right\|_1.
\end{aligned}
\end{equation}

\section{Experiment setup details}
\label{app-sec:exp-setup-details}

\subsection{Details on data sets}
\label{app-sec:exp-setup-details-datasets}

The information of the data sets for OoD detection is outlined below.
All these settings follow previous works.

For the CIFAR10 benchmark, the InD data is the training and test sets of CIFAR10, with 50,000 and 10,000 $32\times32\times3$ images of 10 categories, respectively. In this experiment, all images from the OoD data sets are resized to an image size of $32\times32\times3$.
The evaluated OoD data sets are introduced below.
\begin{itemize}
    \item SVHN \cite{netzer2011reading} data set includes images of street view house numbers.
    The test set of SVHN is adopted for OoD detection with 26,032 digits of numbers 0-9.
    \item LSUN \cite{yu2015lsun} data set is about large-scale scene classification and understanding. The test set of LSUN is employed for OoD detection with 10,000 images of 10 categories.
    \item iSUN \cite{xu2015turkergaze} data set consists of 8,925 images of gaze traces, all of which are employed for OoD detection.
    \item Textures \cite{cimpoi2014describing} data set covers various types of surface texture with 5,640 images of 47 categories. The whole data set is adopted in the evaluation of OoD detection performance.
    \item Places365 \cite{zhou2017places} data set is for scene recognition. A subset of 328,500 images are curated for OoD detection by \cite{sun2022out}.
\end{itemize}

For the ImageNet-1K benchmark, the InD data is the training and test sets of ImageNet-1K, with 1,281,167 and 50,000 images of 1,000 categories, respectively. In experiments, all images from the OoD data sets are resized to an image size of $224\times224\times3$.
The evaluated OoD data sets are introduced below.
\begin{itemize}
    \item iNaturalist \cite{van2018inaturalist} data set contains natural fine-grained images of different species of plants and animals. For OoD detection, 10,000 images are sampled from the selected concepts by \cite{sun2022out}.
    \item SUN \cite{xiao2010sun} data set covers a large variety of environmental scenes, places and the objects within. For OoD detection, 10,000 images are sampled from the selected concepts by \cite{sun2022out}.
    \item Places \cite{zhou2017places} data set is about scene recognition. For OoD detection, 10,000 images are sampled by \cite{sun2022out}.
    \item Textures \cite{cimpoi2014describing} data set in the ImageNet-1K benchmark is the same as that described above in the CIFAR10 benchmark.
\end{itemize}

\begin{table*}[ht]
    \centering
    \begin{tabular}{@{}c|c|c|cc@{}}
    \toprule
    data set & network & \# GPUs & time per Epoch {[s]} & total time {[h]} \\
    \midrule
    \multirow{2}{*}{CIFAR10} & ResNet18 & 3090 $\times1$ & 18.22 & 0.76 \\
    & Wide ResNet28X10 & 3090 $\times1$ & 97.04 & 4.04 \\
    \midrule
    \multirow{3}{*}{ImageNet-1K} & ResNet50 & v100 $\times4$ & 1950.41 & 48.76\\
    & DenseNet121 & v100 $\times4$ & 2439.42 & 60.99 \\
    & T2T-ViT-14 & v100 $\times8$ & 1607.98 & 138.47 \\
    \bottomrule
    \end{tabular}
    \caption{Training time consumed by each individual model with specific GPUs in the experiments.}
    \label{app-tab:training-time}
\end{table*}

\subsection{Details on model training}
\label{app-sec:exp-setup-details-training}

Thorough training details of the adopted modes for OoD detection are elaborated below.
Particularly, to obtain multiple independent modes, it is required to re-train multiple models from scratch on the training sets of CIFAR10 \cite{krizhevsky2009learning} and ImageNet-1K \cite{deng2009imagenet} {\it w.r.t} different random seeds.
For checkable reproducibility of the results reported in this paper, all the training and evaluation code and the trained checkpoints can be found in the publicly-released link given in the main text.

For the 10 independent modes of ResNet18 \cite{he2016deep} and Wide ResNet28X10 \cite{zagoruyko2016wide} trained on CIFAR10 \cite{krizhevsky2009learning}, each DNN is optimized via SGD for 150 epochs with a batch size of 256 and weight decay $10^{-4}$.
The initial learning rate is 0.1 and reduced via a cosine scheduler to 0 during training.
Each DNN is trained on one single NVIDIA GeForce RTX 3090 GPU.

For the 5 independent modes of ResNet50 \cite{he2016deep} and DenseNet121 \cite{huang2017densely} trained from scratch on ImageNet-1K \cite{deng2009imagenet}, we follow the official training script\footnote{\href{https://github.com/pytorch/examples/tree/main/imagenet}{https://github.com/pytorch/examples/tree/main/imagenet}} provided by PyTorch.
Each DNN is optimized via SGD for 90 epochs with weight decay $10^{-4}$.
The initial learning rate is 0.1 and reduced every 30 epochs by a factor of 10.
The batch size for training ResNet50 and DenseNet121 is 1000 and 800, respectively.
Each training is executed parallely on 4 NVIDIA v100 GPUs.

For the 3 independent modes of T2T-ViT-14 \cite{yuan2021tokens} trained from scratch on ImageNet, we follow the training script provided from the official github repository\footnote{\href{https://github.com/yitu-opensource/T2T-ViT}{https://github.com/yitu-opensource/T2T-ViT}} and adopt the default recommendation settings.
Each T2T-ViT-14 model is trained parallely on 8 NVIDIA v100 GPUs for 310 epochs with a batch size of 64, an initial learning rate of $5\times10^{-4}$ and weight decay 0.05.

The final models after the training finishes are evaluated for OoD detection in inference.

\paragraph{Computational overhead}
For training, general ensemble methods inevitably requires extra time and memory expenses in the employed multiple models than that of single-model-based methods \cite{horvath2021boosting,wortsman2022model}.
Similarly in our proposed mode ensemble, multiple modes require training multiple DNNs. 
In Table \ref{app-tab:training-time}, we provide the time-consuming of training a single model, where our $N$-mode ensemble takes such training  $N$ times. 
In implementation, we can proceed the training in parallel to reduce the heavy time cost of training sequentially, given with sufficient computation resources.
In this way, multiple DNNs could be obtained in the time of training one DNN.
Once the models are trained, our OoD detectors are available at hands.
Then, they can be used to detect any given data, where only the inference is needed.

\section{Theoretical proofs}
\label{app-sec:theory-proof}

This section gives the proof of Proposition \ref{prop:gap} in the manuscript, which is reiterated as follows.
\begin{propref}{prop:gap}

Consider the in-distribution ${\cal P}_{\rm in}$ and the out-distribution ${\cal P}_{\rm out}$, and the $N$ modes $g_i,i=1,\cdots,N$, we have $\mathcal{G}(g_{\rm ens},\mathcal{P}_{\rm in},\mathcal{P}_{\rm out})\leq\frac{1}{N}\sum_{i=1}^N\mathcal{G}(g_i,\mathcal{P}_{\rm in},\mathcal{P}_{\rm out})$, where
\begin{equation}
\begin{aligned}
\label{eq:acc-gap-app}
\mathcal{G}(g,\mathcal{P}_{\rm in},\mathcal{P}_{\rm out})=\left|\Phi^{-1}(\mathcal{ACC}(g,\mathcal{P}_{\rm out}))-\frac{\alpha}{\gamma}\Phi^{-1}(\mathcal{ACC}(g,\mathcal{P}_{\rm in}))\right|
\end{aligned}
\end{equation}
denotes the gap of the probit-transformed accuracy between $\mathcal{P}_{\rm in}$ and $\mathcal{P}_{\rm out}$ achieved by the classifier $g$.
\end{propref}

\begin{proof}
Given the in-distribution $\mathcal{P}_{\rm in}$ defined as a Gaussian distribution $\mathcal{P}_{\rm in}=\{(\boldsymbol{x},y)\ |\ \boldsymbol{x}\sim\mathcal{N}(\boldsymbol{\mu}\cdot y;\sigma^2I_{D\times D})\}$ with $\boldsymbol{x}\in\mathbb{R}^D$ and $y\in\{-1,1\}$, and one single mode in the form of a linear classifier $g_i:\boldsymbol{x}\rightarrow\mathrm{sign}(\boldsymbol{\theta}_i^T\boldsymbol{x}),i=1,\cdots,N$, we have
\begin{equation}
\begin{aligned}
\label{eq:acc-gi-in}
\mathcal{ACC}(g_i,\mathcal{P}_{\rm in})
&={\rm Pr}\left(\mathrm{sign}(\boldsymbol{\theta}_i^T\boldsymbol{x})=y\right)={\rm Pr}\left(y\cdot(\boldsymbol{\theta}_i^T\boldsymbol{x})>0\right)\\
&={\rm Pr}\left(\mathcal{N}\left(\boldsymbol{\theta}_i^T\boldsymbol{\mu};\|\boldsymbol{\theta}_i\|^2\sigma^2\right)\geq0\right)
={\rm Pr}\left(\|\boldsymbol{\theta}_i\|\sigma\cdot\mathcal{N}(0;1)\geq-(\boldsymbol{\theta}_i^T\boldsymbol{\mu})\right)
=\Phi\left(\frac{\boldsymbol{\theta}_i^T\boldsymbol{\mu}}{\|\boldsymbol{\theta}_i\|\sigma}\right).
\end{aligned}
\end{equation}
Similarly, for the out-distribution ${\cal{P}}_{\rm out}$ defined as a shift from $\mathcal{P}_{\rm in}$: $\mathcal{P}_{\rm out}=\{(\boldsymbol{x},y)\ |\ \boldsymbol{x}\sim\mathcal{N}(\boldsymbol{\mu}^\prime\cdot y;{\sigma^\prime}^2I_{D\times D})\}$ with $\boldsymbol{\mu}^\prime=\alpha\cdot\boldsymbol{\mu}+\beta\cdot\boldsymbol{\Delta}$ and $\sigma^\prime=\gamma\cdot\sigma$, and the mode ensemble function $g_{\rm ens}:\boldsymbol{x}\rightarrow\mathrm{sign}(\sum_{i=1}^{N}(\boldsymbol{\theta}^T_i\boldsymbol{x}))$, we have
\begin{equation}
\begin{aligned}
\label{eq:acc-other}
&\mathcal{ACC}(g_i,\mathcal{P}_{\rm out})
=\Phi\left(\frac{\boldsymbol{\theta}_i^T\boldsymbol{\mu}^\prime}{\|\boldsymbol{\theta}_i\|\sigma^\prime}\right)
=\Phi\left(\frac{\alpha}{\gamma}\cdot\frac{\boldsymbol{\theta}_i^T\boldsymbol{\mu}}{\|\boldsymbol{\theta}_i\|\sigma}
+\frac{\beta}{\gamma\sigma}\cdot\frac{\boldsymbol{\theta}_i^T\boldsymbol{\Delta}}{\|\boldsymbol{\theta}_i\|}\right),\\
&\mathcal{ACC}(g_{\rm ens},\mathcal{P}_{\rm in})
=\Phi\left(\frac{\sum_{i=1}^N(\boldsymbol{\theta}_i^T\boldsymbol{\mu})}{\|\sum_{i=1}^N\boldsymbol{\theta}_i\|\sigma}\right),\\
&\mathcal{ACC}(g_{\rm ens},\mathcal{P}_{\rm out})
=\Phi\left(\frac{\alpha}{\gamma}\cdot\frac{\sum_{i=1}^N(\boldsymbol{\theta}_i^T\boldsymbol{\mu})}{\|\sum_{i=1}^N\boldsymbol{\theta}_i\|\sigma}
+\frac{\beta}{\gamma\sigma}\cdot\frac{\sum_{i=1}^N(\boldsymbol{\theta}_i^T\boldsymbol{\Delta})}{\|\sum_{i=1}^N\boldsymbol{\theta}_i\|}\right).
\end{aligned}
\end{equation}
Substituting Eq.\eqref{eq:acc-gi-in} and Eq.\eqref{eq:acc-other} into Eq.\eqref{eq:acc-gap-app}, we have
\begin{equation}
\begin{aligned}
\label{eq:acc-gap-all}
&\mathcal{G}_{\rm avg}
\triangleq\frac{1}{N}\sum_{i=1}^N\mathcal{G}(g_i,{\cal P_{\rm in}},{\cal P_{\rm out}})
=\frac{\beta}{\gamma\sigma}\cdot\frac{1}{N}\sum_{i=1}^N\frac{|\boldsymbol{\theta}_i^T\boldsymbol{\Delta}|}{\|\boldsymbol{\theta}_i\|},\\
&\mathcal{G}_{\rm ens}
\triangleq\mathcal{G}(g_{\rm ens},\mathcal{P}_{\rm in},\mathcal{P}_{\rm out})
=\frac{\beta}{\gamma\sigma}\cdot\frac{|\sum_{i=1}^N(\boldsymbol{\theta}_i^T\boldsymbol{\Delta})|}{\|\sum_{i=1}^N\boldsymbol{\theta}_i\|}.
\end{aligned}
\end{equation}
If we are able to show that ${\cal G}_{\rm ens}\leq{\cal G}_{\rm avg}$ then the proof is finished.
The derivations continue as follows.
\begin{equation}
\begin{aligned}
\label{eq:acc-gap-all-reduce}
&\mathcal{G}_{\rm avg}
=\frac{\beta}{\gamma\sigma}\cdot\frac{1}{N}\sum_{i=1}^N\frac{\|\boldsymbol{\theta}_i\|\|\boldsymbol{\Delta}\||\cos\langle\boldsymbol{\theta}_i,\boldsymbol{\Delta}\rangle|}{\|\boldsymbol{\theta}_i\|}
=\frac{\beta}{\gamma\sigma}\cdot\|\boldsymbol{\Delta}\|\frac{|\cos\langle\boldsymbol{\theta}_1,\boldsymbol{\Delta}\rangle|+\cdots+|\cos\langle\boldsymbol{\theta}_N,\boldsymbol{\Delta}\rangle|}{N},\\
&\mathcal{G}_{\rm ens}
=\frac{\beta}{\gamma\sigma}\cdot\frac{|\sum_{i=1}^N(\|\boldsymbol{\theta}_i\|\|\boldsymbol{\Delta}\|\cos\langle\boldsymbol{\theta}_i,\boldsymbol{\Delta}\rangle)|}{\|\sum_{i=1}^N\boldsymbol{\theta}_i\|}
=\frac{\beta}{\gamma\sigma}\cdot\|\boldsymbol{\Delta}\|\left|\frac{\|\boldsymbol{\theta}_1\|\cos\langle\boldsymbol{\theta}_1,\boldsymbol{\Delta}\rangle+\cdots+\|\boldsymbol{\theta}_N\|\cos\langle\boldsymbol{\theta}_N,\boldsymbol{\Delta}\rangle}{\|\boldsymbol{\theta}_1+\cdots+\boldsymbol{\theta}_N\|}\right|,\\
\end{aligned}
\end{equation}
Now we make 2 assumptions:
\begin{enumerate}
\item The parameters $\boldsymbol{\theta}_i$ of these modes hold similar norms: $\|\boldsymbol{\theta}_1\|\approx\|\boldsymbol{\theta}_2\|\approx\cdots\approx\|\boldsymbol{\theta}_N\|$.
\item The parameters $\boldsymbol{\theta}_i$ of these modes are along one nearly consistent direction, \textit{i.e.}, the angle between any 2 different parameters $\boldsymbol{\theta}_i$ and $\boldsymbol{\theta}_j$ is nearly 0.
\end{enumerate}
Then,
\begin{equation}
\begin{aligned}
\label{eq:gap-division}
\frac{\cal{G}_{\rm avg}}{\cal{G}_{\rm ens}}
&=\frac{|\cos\langle\boldsymbol{\theta}_1,\boldsymbol{\Delta}\rangle|+\cdots+|\cos\langle\boldsymbol{\theta}_N,\boldsymbol{\Delta}\rangle|}{|\|\boldsymbol{\theta}_1\|\cos\langle\boldsymbol{\theta}_1,\boldsymbol{\Delta}\rangle+\cdots+\|\boldsymbol{\theta}_N\|\cos\langle\boldsymbol{\theta}_N,\boldsymbol{\Delta}\rangle|}\cdot\frac{\|\boldsymbol{\theta}_1+\cdots+\boldsymbol{\theta}_N\|}{N}\\
&=\frac{|\cos\langle\boldsymbol{\theta}_1,\boldsymbol{\Delta}\rangle|+\cdots+|\cos\langle\boldsymbol{\theta}_N,\boldsymbol{\Delta}\rangle|}{|\cos\langle\boldsymbol{\theta}_1,\boldsymbol{\Delta}\rangle+\cdots+\cos\langle\boldsymbol{\theta}_N,\boldsymbol{\Delta}\rangle|}\geq 1.
\end{aligned}
\end{equation}
The Proposition \ref{prop:gap} is proved.
\end{proof}

\section{Comprehensive empirical results}
\label{app-sec:all-exp}

In this section, the comprehensive empirical results on OoD detection covering all the detectors, network structures and data sets are provided.
The recognition results on the InD test data of all the modes trained on CIFAR10 and ImageNet-1K are shown in Sec.\ref{app-sec:exp-classification}.
Section \ref{app-sec:exp-ood-single} and Section \ref{app-sec:exp-ood-ensemble} illustrate the complete OoD detection results of each single mode and mode ensemble, respectively.

\subsection{Classification accuracy on InD data}
\label{app-sec:exp-classification}

The classification accuracy of all the independent modes on their InD data sets CIFAR10 and ImageNet-1K is presented in Tab.\ref{app-tab:exp-indacc-c10} and Tab.\ref{app-tab:exp-indacc-imgnet}, respectively.
It is clearly shown that these isolated modes all hold consistently good recognition performance on their corresponding InD test sets.

\begin{table}[h]
    \centering
    \resizebox{\hsize}{!}{
    \begin{tabular}{@{}cc|ccccc ccccc@{}}
    \toprule
    network & capacity & mode-1 & mode-2 & mode-3 & mode-4 & mode-5 & mode-6 & mode-7 & mode-8 & mode-9 & mode-10 \\
    \midrule
    RN18 & 11,173,962 & 93.21 & 93.33 & 93.61 & 93.26 & 93.68 & 93.57 & 93.60 & 93.85 & 93.79 & 93.91 \\
    WRN28X10 & 36,489,290 & 94.02 & 94.30 & 93.54 & 94.12 & 94.29 & 93.80 & 94.39 & 94.01 & 94.27 & 94.03 \\
    \bottomrule
    \end{tabular}}
    \caption{The classification accuracy (\%) of 10 isolated modes of 2 network structures, \textbf{RN18} and \textbf{WRN28X10}, on the \textbf{CIFAR10} test set as the InD data.}
    \label{app-tab:exp-indacc-c10}
\end{table}

\begin{table}[h]
    \centering
    \begin{tabular}{@{}cc|ccccc @{}}
    \toprule
    network & capacity & mode-1 & mode-2 & mode-3 & mode-4 & mode-5 \\
    \midrule
    RN50 & 25,557,032 & 74.84 & 74.71 & 74.86 & 74.65 & 74.80 \\
    DN121 & 7,978,856 & 74.25 & 74.30 & 74.46 & 74.29 & 74.33 \\
    \midrule
    T2T-ViT-14 & 21,545,550 & 81.30 & 81.44 & 81.53 & - & - \\
    \bottomrule
    \end{tabular}
    \caption{The classification accuracy (\%) of 5 isolated modes of 2 network structures, \textbf{RN50} and \textbf{DN121}, and 3 isolated modes of \textbf{T2T-ViT-14}, on the \textbf{ImageNet-1K} validation set as the InD data.}
    \label{app-tab:exp-indacc-imgnet}
\end{table}

\subsection{OoD detection results of single modes}
\label{app-sec:exp-ood-single}
In this section, we provide the complete detection results on the OoD data sets of all single modes of different network structures trained on the CIFAR10 and ImageNet-1K benchmarks \textit{w.r.t} different types of detectors.

\paragraph{Results on CIFAR10.} 
In Tab.\ref{app-tab:exp-uncertainty-c10-r18} and Tab.\ref{app-tab:exp-uncertainty-c10-wrn28x10}, we record the detection performance of 10 isolated modes trained on CIFAR10 of ResNet18 and Wide ResNet28X10, respectively.
The \underline{\bf best} and \uwave{\bf worst} detection results are marked in the 2 tables for each mode on each OoD data set \textit{w.r.t} 5 OoD detectors: MSP, ODIN, Energy, Mahalanobis and KNN.

As shown in Tab.\ref{app-tab:exp-uncertainty-c10-r18} and Tab.\ref{app-tab:exp-uncertainty-c10-wrn28x10}, those isolated and independent modes clearly hold significantly varied OoD detection results on multiple OoD data sets for different types of OoD detectors.
Taking the MSP method of ResNet18 as an example, the FPR values on the SVHN data set could range from the worst 77.76\% to the best 65.20\% with a 12.56\% difference among the 10 modes.
Besides, considering the mode-4 of ResNet18, ODIN achieves lower FPR values on SVHN (55.85\%) than MSP (66.45\%), while for mode-9, MSP (67.52\%) instead outperforms ODIN (72.09\%). 
The results of other OoD detectors on other OoD data sets all manifest such a common phenomenon, \textit{i.e.}, a high variance on the OoD detection performance among the isolated and independent modes, which further indicates the high uncertainty of the detection performance on OoD data for single modes.

\paragraph{Results on ImageNet-1K.}
In Tab.\ref{app-tab:exp-uncertainty-imgnet-r50} and Tab.\ref{app-tab:exp-uncertainty-imgnet-dn121}, we record the detection performance of 5 isolated modes trained from scratch on ImageNet-1K of ResNet50 and DenseNet121, respectively.
The \underline{\bf best} and \uwave{\bf worst} detection results are marked in the 2 tables for each mode on each OoD data set \textit{w.r.t} 6 OoD detectors: MSP, ODIN, Energy, GradNorm, Mahalanobis and RankFeat.

As shown in Tab.\ref{app-tab:exp-uncertainty-imgnet-r50} and Tab.\ref{app-tab:exp-uncertainty-imgnet-dn121}, compared to the results on CIFAR10, those isolated and independent modes shows relatively more consistent detection performance.
This might be due to that the training set size of ImageNet-1K (1,281,167) is much more larger than that of CIFAR10 (50,000), which benefits the model generalization both on InD data and OoD data as the model has ``seen'' much more numerous images during training.
Nevertheless, the detection variances among those modes cannot be simply ignored, and the variance of DenseNet121 is larger and more evident than that of ResNet50, indicating that modes of different structures show different uncertainties.
We reiterate the 2 phenomena that reflect the high detection variances existed in the independent modes on both CIFAR10 and ImageNet-1K, which has been summarized in the manuscript:
\begin{itemize}
    \item For a given OoD detection method, those independent modes hold significantly-fluctuating FPR results.
    \item For different modes, detection results among those detection methods can also be opposite.
    That is, detector A might outperform detector B on some mode, while detector B shows superiority over detector A on another mode.
\end{itemize}

\subsection{OoD detection results of mode ensemble}
\label{app-sec:exp-ood-ensemble}
In this section, we provide the complete detection results on the OoD data sets of ensembling multiple independent modes of different network structures on CIFAR10 and ImageNet-1K benchmarks \textit{w.r.t} different types of detectors.
The details of ensembling ways for each OoD detector have been elaborated in Appendix A of the main text.

\paragraph{Results on CIFAR10.} 
For the 10 independent modes on CIFAR10, we ensemble $k\in\{1,2,4,6,8\}$ modes to verify the effectiveness of mode ensemble for OoD detection, shown in Tab.\ref{app-tab:exp-ensemble-c10-r18} and Tab.\ref{app-tab:exp-ensemble-c10-wrn28x10} for ResNet18 and Wide ResNet28X10, respectively.
For $k=1$, we report the average detection results of the total 10 modes with the mean values and standard deviations.
For $k\in\{2,4,6,8\}$, we randomly pick out $k$ \textit{different} modes from the 10 modes and perform mode ensemble, which is repeated 3 times, then the average detection results of the 3 random pickings are presented with the mean values and standard deviations.

As shown in Tab.\ref{app-tab:exp-ensemble-c10-r18} and Tab.\ref{app-tab:exp-ensemble-c10-wrn28x10}, by ensembling multiple isolated modes, the variances of the detection results get reduced and the detection performance gets significantly improved.
Taking the Mahalanobis method with ResNet18 as an example, its average detection FPR on SVHN of single modes is up to 24.66\%, which is improved to 3.81\% with a 20.85\% reduction by ensembling 8 modes, while the variance simultaneously gets reduced from 12.72\% to 0.52\%.

\paragraph{Results on ImageNet-1K.}
For the 5 independent modes on ImageNet-1K, we ensemble $k\in\{1,2,3,4\}$ modes to verify the effectiveness of mode ensemble for OoD detection, shown in Tab.\ref{app-tab:exp-ensemble-imgnet-r50} and Tab.\ref{app-tab:exp-ensemble-imgnet-dn121} for ResNet50 and DenseNet121, respectively.
For $k=1$, we report the average detection results of the total 5 modes with the mean values and standard deviations.
For $k\in\{2,3,4\}$, we randomly pick out $k$ \textit{different} modes from the 5 modes and perform mode ensemble, which is repeated 3 times, then the average detection results of the 3 random pickings are presented with the mean values and standard deviations.

As shown in Tab.\ref{app-tab:exp-ensemble-imgnet-r50} and Tab.\ref{app-tab:exp-ensemble-imgnet-dn121}, the mode ensemble could bring variance reduction and performance improvements for different OoD detectors.
For example, considering the RankFeat method with DenseNet121, the average and the standard deviation of the detection FPR values on iNaturalist could be improved from 63.06\% to 40.92\% with a 22.14\% reduction and from 13.98\% to 5.41\% with an 8.57\% reduction, respectively.

In short, these detection results on both CIFAR10 and ImageNet-1K benchmarks with various OoD detectors and network structures have shown the 2 benefits brought by mode ensemble:
\begin{itemize}
    \item For a given OoD detection method, its performance gets significantly improved via mode ensemble with both reduced variances and better detection results.
    \item Mode ensemble benefits the evaluation among different OoD detectors as it gets rid of the uncertainties of single modes.
\end{itemize}

\end{appendices}

\clearpage

{\small
\begin{longtable}[t]{H c cc cc cc cc cc HH}
\caption{The detection performance of each mode of {\bf ResNet18} trained on {\bf CIFAR10} \textit{w.r.t} 5 OoD detectors: MSP, ODIN, Energy, Mahalanobis and kNN. The \underline{\bf best} and \uwave{\bf worst} results of each mode on each OoD data set are highlighted.}
\label{app-tab:exp-uncertainty-c10-r18}\\
\toprule
\multirow{3}{*}{Network} & \multirow{3}{*}{Modes}  & \multicolumn{12}{c}{OoD data sets} \\
& & \multicolumn{2}{c}{SVHN} & \multicolumn{2}{c}{LSUN} & \multicolumn{2}{c}{iSUN} & \multicolumn{2}{c}{Texture} & \multicolumn{2}{c}{Places365} & & \\
& & FPR$\downarrow$ & AUROC$\uparrow$ & FPR$\downarrow$ & AUROC$\uparrow$ & FPR$\downarrow$ & AUROC$\uparrow$ & FPR$\downarrow$ & AUROC$\uparrow$ & FPR$\downarrow$ & AUROC$\uparrow$ & FPR$\downarrow$ & AUROC$\uparrow$ \\
\midrule
\endfirsthead

\multicolumn{14}{c}{{\bfseries \tablename\ \thetable{} -- continued from previous page}} \\
\toprule
\multirow{3}{*}{Network} & \multirow{3}{*}{Modes}  & \multicolumn{12}{c}{OoD data sets} \\
& & \multicolumn{2}{c}{SVHN} & \multicolumn{2}{c}{LSUN} & \multicolumn{2}{c}{iSUN} & \multicolumn{2}{c}{Texture} & \multicolumn{2}{c}{Places365} & & \\
& & FPR$\downarrow$ & AUROC$\uparrow$ & FPR$\downarrow$ & AUROC$\uparrow$ & FPR$\downarrow$ & AUROC$\uparrow$ & FPR$\downarrow$ & AUROC$\uparrow$ & FPR$\downarrow$ & AUROC$\uparrow$ & FPR$\downarrow$ & AUROC$\uparrow$ \\
\midrule
\endhead

\midrule
\multicolumn{14}{r}{{Continued on next page}} \\ 
\midrule
\endfoot

\endlastfoot

\multicolumn{14}{c}{\textbf{MSP}}\\
&mode-1 & 74.87 & 83.87 & 55.00 & 91.70 & 52.67 & 92.09 & \uwave{\bf 67.36} & \uwave{\bf 87.26} & 64.57 & 87.60 \\
&mode-2 & 71.58 & 86.86 & 53.19 & 92.07 & \uwave{\bf 65.85} & 88.92 & 65.07 & 88.25 & 65.23 & 87.76 \\
&mode-3 & 74.87 & 83.69 & \uwave{\textbf{56.30}} & \uwave{\textbf{90.94}} & 63.74 & 90.17 & 64.79 & 88.53 & \uwave{\bf 65.61} & \uwave{\bf 87.46} \\
&mode-4 & 66.45 & 89.15 & 50.01 & 92.52 & 61.85 & 90.21 & 65.66 & 88.63 & 64.97 & 87.84 \\
&mode-5 & 73.32 & 84.69 & 48.30 & 93.19 & 60.86 & 89.45 & 63.10 & 88.60 & 63.14 & 87.78 \\
&mode-6 & \textbf{\underline{65.20}} & 89.09 & 51.25 & 92.44 & 59.88 & 90.84 & \underline{\bf 61.60} & \underline{\bf 89.92} & 63.56 & 87.89 \\
&mode-7 & 77.26 & 86.00 & 50.81 & 92.50 & 65.21 & \uwave{\bf 88.74} & 63.21 & 88.31 & 63.53 & 87.55 \\
\bottomrule
\end{longtable}
}

{\small
\begin{longtable}[t]{H c cc cc cc cc cc HH}
    \caption{The detection performance of each mode of {\bf Wide ResNet28X10} trained on {\bf CIFAR10} \textit{w.r.t} 5 OoD detectors: MSP, ODIN, Energy, Mahalanobis and kNN. The \underline{\bf best} and \uwave{\bf worst} results of each mode are highlighted.}
    \label{app-tab:exp-uncertainty-c10-wrn28x10}\\
    \toprule
    \multirow{3}{*}{Network} & \multirow{3}{*}{Modes}  & \multicolumn{12}{c}{OoD data sets} \\
    & & \multicolumn{2}{c}{SVHN} & \multicolumn{2}{c}{LSUN} & \multicolumn{2}{c}{iSUN} & \multicolumn{2}{c}{Texture} & \multicolumn{2}{c}{Places365} & & \\
    & & FPR$\downarrow$ & AUROC$\uparrow$ & FPR$\downarrow$ & AUROC$\uparrow$ & FPR$\downarrow$ & AUROC$\uparrow$ & FPR$\downarrow$ & AUROC$\uparrow$ & FPR$\downarrow$ & AUROC$\uparrow$ & FPR$\downarrow$ & AUROC$\uparrow$ \\
    \midrule
    \endfirsthead

    \multicolumn{14}{c}{{\bfseries \tablename\ \thetable{} -- continued from previous page}} \\
    \toprule
    \multirow{3}{*}{Network} & \multirow{3}{*}{Modes}  & \multicolumn{12}{c}{OoD data sets} \\
    & & \multicolumn{2}{c}{SVHN} & \multicolumn{2}{c}{LSUN} & \multicolumn{2}{c}{iSUN} & \multicolumn{2}{c}{Texture} & \multicolumn{2}{c}{Places365} & & \\
    & & FPR$\downarrow$ & AUROC$\uparrow$ & FPR$\downarrow$ & AUROC$\uparrow$ & FPR$\downarrow$ & AUROC$\uparrow$ & FPR$\downarrow$ & AUROC$\uparrow$ & FPR$\downarrow$ & AUROC$\uparrow$ & FPR$\downarrow$ & AUROC$\uparrow$ \\
    \midrule
    \endhead

    \midrule
    \multicolumn{14}{r}{{Continued on next page}} \\ 
    \midrule
    \endfoot
    
    \endlastfoot
    
    \multicolumn{14}{c}{\textbf{MSP}}\\
    &mode-1 & 54.52 & 93.54 & \uwave{\bf 32.95} & \uwave{\bf 95.66} & 46.82 & 93.59 & 60.90 & 89.12 & 62.48 & 88.44 \\
    &mode-2 & \underline{\bf 30.10} & 95.62 & 27.41 & 96.29 & 52.13 & 92.71 & \uwave{\bf 64.04} & 88.08 & \underline{\bf 57.17} & \underline{\bf 89.84} \\
    &mode-3 & 45.38 & 94.01 & 23.21 & 96.57 & 47.41 & 93.58 & 53.65 & 90.95 & \uwave{\bf 62.54} & \uwave{\bf 88.12} \\
    &mode-4 & \uwave{\bf 60.18} & \uwave{\bf 91.23} & 32.71 & 95.80 & 52.25 & 92.83 & 62.27 & 89.82 & 62.00 & 88.22 \\
    &mode-5 & 43.60 & 94.00 & 26.34 & 96.25 & 52.25 & 92.77 & 60.55 & 89.30 & 59.62 & 89.06 \\
    &mode-6 & 37.53 & 94.38 & \underline{\bf 17.96} & \underline{\bf 97.26} & 40.64 & 94.23 & \underline{\bf 48.32} & \underline{\bf 91.76} & 59.19 & 88.83 \\
    &mode-7 & 53.86 & 92.12 & 27.23 & 96.13 & \underline{\bf 40.19} & \underline{\bf 94.46} & 60.30 & \uwave{\bf 88.03} & 57.34 & 89.55 \\
\bottomrule
\end{longtable}
}

\begin{table*}[ht]
    \centering
    \begin{tabular}{@{}H c cc cc cc cc HH@{}}
    \toprule
    \multirow{3}{*}{Network} & \multirow{3}{*}{Modes}  & \multicolumn{10}{c}{OoD data sets} \\
    & & \multicolumn{2}{c}{iNaturalist} & \multicolumn{2}{c}{SUN} & \multicolumn{2}{c}{Places} & \multicolumn{2}{c}{Texture} & & \\
    & & FPR$\downarrow$ & AUROC$\uparrow$ & FPR$\downarrow$ & AUROC$\uparrow$ & FPR$\downarrow$ & AUROC$\uparrow$ & FPR$\downarrow$ & AUROC$\uparrow$ & FPR$\downarrow$ & AUROC$\uparrow$ \\
    \midrule
    
    \multicolumn{12}{c}{\textbf{MSP}}\\
    &mode-1 & 57.33 & 87.29 & 70.84 & \uwave{\bf 80.65} & 72.49 & \underline{\bf 80.44} & 67.20 & \underline{\bf 81.10} \\
    &mode-2 & \underline{\bf 55.43} & \underline{\bf 87.95} & 70.50 & \underline{\bf 81.35} & 72.82 & 80.43 & \uwave{\bf 69.54} & 79.88 \\
    &mode-3 & 55.84 & 87.89 & \underline{\bf 70.06} & 81.11 & 72.63 & 80.31 & \underline{\bf 65.89} & 81.04 \\
    &mode-4 & 55.95 & 87.49 & 70.81 & 80.70 & \uwave{\bf 72.96} & \uwave{\bf 80.15} & 68.33 & \uwave{\bf 79.84} \\
\bottomrule
    \end{tabular}
    \caption{The detection performance of each of the 5 modes of {\bf ResNet50} trained from scratch on {\bf ImageNet-1K} \textit{w.r.t} 6 OoD detectors: MSP, ODIN, Energy, GradNorm, Mahalanobis and RankFeat. The \underline{\bf best} and \uwave{\bf worst} results of each mode on each OoD data set are highlighted.}
    \label{app-tab:exp-uncertainty-imgnet-r50}
\end{table*}

\begin{table*}[ht]
    \centering
    \begin{tabular}{@{}H c cc cc cc cc HH@{}}
    \toprule
    \multirow{3}{*}{Network} & \multirow{3}{*}{Modes}  & \multicolumn{10}{c}{OoD data sets} \\
    & & \multicolumn{2}{c}{iNaturalist} & \multicolumn{2}{c}{SUN} & \multicolumn{2}{c}{Places} & \multicolumn{2}{c}{Texture} & & \\
    & & FPR$\downarrow$ & AUROC$\uparrow$ & FPR$\downarrow$ & AUROC$\uparrow$ & FPR$\downarrow$ & AUROC$\uparrow$ & FPR$\downarrow$ & AUROC$\uparrow$ & FPR$\downarrow$ & AUROC$\uparrow$ \\
    \midrule
    
    \multicolumn{12}{c}{\textbf{MSP}}\\
    &mode-1 & 58.44 & 87.18 & \uwave{\bf 71.11} & 80.78 & \uwave{\bf 74.11} & \uwave{\bf 79.69} & 67.07 & 80.72 \\
    &mode-2 & \underline{\bf 55.68} & \underline{\bf 87.28} & \underline{\bf 68.78} & \underline{\bf 81.49} & \underline{\bf 70.85} & \underline{\bf 80.63} & 66.21 & \uwave{\bf 80.27} \\
    &mode-3 & 58.29 & 87.10 & 70.60 & 81.02 & 73.01 & 80.07 & 66.06 & 80.70 \\
    &mode-4 & 58.52 & 87.20 & 70.73 & 80.81 & 72.79 & 79.77 & \uwave{\bf 67.85} & 80.46 \\
\bottomrule
    \end{tabular}
    \caption{The detection performance of each of the 5 modes of {\bf DenseNet121} trained from scratch on {\bf ImageNet-1K} \textit{w.r.t} 6 OoD detectors: MSP, ODIN, Energy, GradNorm, Mahalanobis and RankFeat. The \underline{\bf best} and \uwave{\bf worst} results of each mode on each OoD data set are highlighted.}
    \label{app-tab:exp-uncertainty-imgnet-dn121}
\end{table*}

\begin{table*}[t]
    \centering
    \resizebox{\textwidth}{!}{
    \begin{tabular}{@{}c H cc cc cc cc cc HH@{}}
    \toprule
    \multirow{3}{*}{\makecell[c]{Ensemble\\of $k$ modes}} & \multirow{3}{*}{InD ACC.}  & \multicolumn{12}{c}{OoD data sets} \\
    & & \multicolumn{2}{c}{SVHN} & \multicolumn{2}{c}{LSUN} & \multicolumn{2}{c}{iSUN} & \multicolumn{2}{c}{Texture} & \multicolumn{2}{c}{Places365} & & \\
    & & FPR$\downarrow$ & AUROC$\uparrow$ & FPR$\downarrow$ & AUROC$\uparrow$ &
    FPR$\downarrow$ & AUROC$\uparrow$ & FPR$\downarrow$ & AUROC$\uparrow$ & FPR$\downarrow$ & AUROC$\uparrow$ & FPR$\downarrow$ & AUROC$\uparrow$ \\
    \midrule

    \multicolumn{14}{c}{\textbf{MSP}}\\
    $k=1$ & & 71.68$\pm$4.61 & 86.51$\pm$2.53 & 51.09$\pm$3.66 & 92.49$\pm$0.84 & 59.82$\pm$5.23 & 90.51$\pm$1.34 & 64.19$\pm$1.60 & 88.58$\pm$0.66 & 64.15$\pm$0.98 & 87.77$\pm$0.23 \\
    $k=2$ & & 68.23$\pm$5.74 & 88.46$\pm$1.84 & 43.56$\pm$6.07 & 93.75$\pm$0.71 & 51.71$\pm$4.49 & 92.68$\pm$0.89 & 58.54$\pm$2.26 & 90.35$\pm$0.35 & 59.35$\pm$1.15 & 89.10$\pm$0.14 \\
    $k=4$ & & 66.34$\pm$5.25 & 88.95$\pm$2.60 & 44.98$\pm$3.70 & 93.66$\pm$0.63 & 55.71$\pm$2.39 & 91.88$\pm$0.54 & 57.69$\pm$0.47 & 90.79$\pm$0.33 & 58.87$\pm$0.57 & 89.49$\pm$0.14 \\
    $k=6$ & & \cellcolor{tabgray}{\bf 67.03$\pm$0.91} & \cellcolor{tabgray}{\bf 89.34$\pm$0.68} & \cellcolor{tabgray}{\bf 44.99$\pm$1.40} & \cellcolor{tabgray}{\bf 93.82$\pm$0.21} & 55.04$\pm$2.61 & 92.31$\pm$0.76 & \cellcolor{tabgray}{\bf 57.49$\pm$0.35} & \cellcolor{tabgray}{\bf 91.03$\pm$0.09} & \cellcolor{tabgray}{\bf 58.75$\pm$0.37} & \cellcolor{tabgray}{\bf 89.73$\pm$0.05} \\
\bottomrule
    \end{tabular}}
    \caption{The detection performance of ensembling $k$ modes (\textbf{ResNet18} trained on \textbf{CIFAR10}) on each OoD data set \textit{w.r.t} different types of OoD detectors. The results with the {\bf lowest variances} are highlighted with bold fonts.}
    \label{app-tab:exp-ensemble-c10-r18}
\end{table*}

\begin{table*}[t]
    \centering
    \resizebox{\textwidth}{!}{
    \begin{tabular}{@{}c H cc cc cc cc cc HH@{}}
    \toprule
    \multirow{3}{*}{\makecell[c]{Ensemble\\of $k$ modes}} & \multirow{3}{*}{InD ACC.}  & \multicolumn{12}{c}{OoD data sets} \\
    & & \multicolumn{2}{c}{SVHN} & \multicolumn{2}{c}{LSUN} & \multicolumn{2}{c}{iSUN} & \multicolumn{2}{c}{Texture} & \multicolumn{2}{c}{Places365} & & \\
    & & FPR$\downarrow$ & AUROC$\uparrow$ & FPR$\downarrow$ & AUROC$\uparrow$ &
    FPR$\downarrow$ & AUROC$\uparrow$ & FPR$\downarrow$ & AUROC$\uparrow$ & FPR$\downarrow$ & AUROC$\uparrow$ & FPR$\downarrow$ & AUROC$\uparrow$ \\
    \midrule

    \multicolumn{14}{c}{\textbf{MSP}}\\
    $k=1$ & & 44.79$\pm$9.78 & 93.87$\pm$1.37 & 26.32$\pm$4.66 & 96.32$\pm$0.48 & 49.01$\pm$7.75 & 93.11$\pm$1.32 & 58.35$\pm$5.00 & 89.75$\pm$1.35 & 59.82$\pm$2.04 & 88.93$\pm$0.62 \\
    $k=2$ & & 33.72$\pm$4.01 & 95.36$\pm$0.42 & 20.88$\pm$1.48 & 96.92$\pm$0.13 & 45.66$\pm$5.05 & 93.92$\pm$0.74 & 54.54$\pm$2.22 & 90.77$\pm$0.74 & 55.88$\pm$1.58 & 89.99$\pm$0.40 \\
    $k=4$ & & \cellcolor{tabgray}{\bf 35.16$\pm$1.04} & 95.17$\pm$0.30 & 18.68$\pm$1.53 & \cellcolor{tabgray}{\bf 97.11$\pm$0.08} & \cellcolor{tabgray}{\bf 36.86$\pm$2.28} & \cellcolor{tabgray}{\bf 95.13$\pm$0.16} & \cellcolor{tabgray}{\bf 52.10$\pm$1.14} & 91.43$\pm$0.41 & 54.04$\pm$1.46 & 90.50$\pm$0.21 \\
    $k=6$ & & 36.91$\pm$2.24 & 95.23$\pm$0.24 & 20.10$\pm$1.15 & 97.04$\pm$0.16 & 40.51$\pm$5.81 & 94.88$\pm$0.62 & 53.21$\pm$2.89 & 91.65$\pm$0.56 & \cellcolor{tabgray}{\bf 55.11$\pm$0.42} & \cellcolor{tabgray}{\bf 90.50$\pm$0.03} \\
\bottomrule
    \end{tabular}}
    \caption{The detection performance of ensembling $k$ modes (\textbf{Wide ResNet28X10} trained on \textbf{CIFAR10}) on each OoD data set \textit{w.r.t} different types of OoD detectors. The results with the {\bf lowest variances} are highlighted with bold fonts.}
    \label{app-tab:exp-ensemble-c10-wrn28x10}
\end{table*}

\begin{table*}[t]
    \centering
    \resizebox{\textwidth}{!}{
    \begin{tabular}{@{}c H cc cc cc cc HH@{}}
    \toprule
    \multirow{3}{*}{\makecell[c]{Ensemble\\of $k$ modes}} & \multirow{3}{*}{InD ACC.}  & \multicolumn{10}{c}{OoD data sets} \\
    & & \multicolumn{2}{c}{iNaturalist} & \multicolumn{2}{c}{SUN} & \multicolumn{2}{c}{Places} & \multicolumn{2}{c}{Textures} & & \\
    & & FPR$\downarrow$ & AUROC$\uparrow$ & FPR$\downarrow$ & AUROC$\uparrow$ &
    FPR$\downarrow$ & AUROC$\uparrow$ & FPR$\downarrow$ & AUROC$\uparrow$ & FPR$\downarrow$ & AUROC$\uparrow$ \\
    \midrule

    \multicolumn{12}{c}{\textbf{MSP}}\\
    $k=1$ & & 56.86$\pm$1.76 & 87.50$\pm$0.45 & 70.75$\pm$0.55 & 80.92$\pm$0.30 & 72.65$\pm$0.25 & 80.32$\pm$0.12 & 68.01$\pm$1.48 & 80.45$\pm$0.61 \\
    $k=2$ & & 53.08$\pm$1.10 & 88.64$\pm$0.29 & 69.41$\pm$0.39 & 81.71$\pm$0.12 & 70.79$\pm$0.22 & 81.18$\pm$0.11 & 65.46$\pm$0.63 & 81.44$\pm$0.26 \\
    $k=3$ & & 52.22$\pm$0.68 & 88.98$\pm$0.21 & 69.09$\pm$0.33 & 82.00$\pm$0.09 & \cellcolor{tabgray}{\bf 70.74$\pm$0.18} & 81.42$\pm$0.03 & 64.74$\pm$1.26 & 81.82$\pm$0.39 \\
    $k=4$ & & \cellcolor{tabgray}{\bf 51.78$\pm$0.55} & \cellcolor{tabgray}{\bf 89.09$\pm$0.06} & \cellcolor{tabgray}{\bf 69.02$\pm$0.10} & \cellcolor{tabgray}{\bf 82.08$\pm$0.07} & 70.24$\pm$0.24 & \cellcolor{tabgray}{\bf 81.58$\pm$0.02} & \cellcolor{tabgray}{\bf 63.70$\pm$0.66} & \cellcolor{tabgray}{\bf 82.10$\pm$0.20} \\
\bottomrule
    \end{tabular}}
    \caption{The detection performance of ensembling $k$ modes (\textbf{ResNet50} trained on \textbf{ImageNet-1K}) on each OoD data set \textit{w.r.t} different types of OoD detectors. The results with the {\bf lowest variances} are highlighted with bold fonts.}
    \label{app-tab:exp-ensemble-imgnet-r50}
\end{table*}

\begin{table*}[t]
    \centering
    \resizebox{\textwidth}{!}{
    \begin{tabular}{@{}c H cc cc cc cc HH@{}}
    \toprule
    \multirow{3}{*}{\makecell[c]{Ensemble\\of $k$ modes}} & \multirow{3}{*}{InD ACC.}  & \multicolumn{10}{c}{OoD data sets} \\
    & & \multicolumn{2}{c}{iNaturalist} & \multicolumn{2}{c}{SUN} & \multicolumn{2}{c}{Places} & \multicolumn{2}{c}{Textures} & & \\
    & & FPR$\downarrow$ & AUROC$\uparrow$ & FPR$\downarrow$ & AUROC$\uparrow$ &
    FPR$\downarrow$ & AUROC$\uparrow$ & FPR$\downarrow$ & AUROC$\uparrow$ & FPR$\downarrow$ & AUROC$\uparrow$ \\
    \midrule

    \multicolumn{12}{c}{\textbf{MSP}}\\
    $k=1$ & & 58.38$\pm$1.87 & 86.99$\pm$0.45 & 70.41$\pm$0.93 & 80.96$\pm$0.32 & 72.70$\pm$1.17 & 80.01$\pm$0.37 & 66.32$\pm$1.29 & 80.62$\pm$0.26 \\
    $k=2$ & & 54.69$\pm$1.01 & 88.29$\pm$0.24 & 68.75$\pm$0.39 & 82.03$\pm$0.04 & 70.72$\pm$0.37 & 81.21$\pm$0.12 & 64.05$\pm$0.57 & 81.58$\pm$0.05 \\
    $k=3$ & & 53.57$\pm$0.55 & 88.75$\pm$0.19 & 68.31$\pm$0.21 & 82.18$\pm$0.09 & 70.74$\pm$0.37 & 81.38$\pm$0.12 & 63.39$\pm$0.48 & \cellcolor{tabgray}{\bf 81.99$\pm$0.02} \\
    $k=4$ & & \cellcolor{tabgray}{\bf 53.42$\pm$0.49} & \cellcolor{tabgray}{\bf 88.94$\pm$0.11} & \cellcolor{tabgray}{\bf 68.48$\pm$0.15} & \cellcolor{tabgray}{\bf 82.34$\pm$0.01} & \cellcolor{tabgray}{\bf 70.47$\pm$0.35} & \cellcolor{tabgray}{\bf 81.52$\pm$0.06} & \cellcolor{tabgray}{\bf 63.35$\pm$0.78} & 82.11$\pm$0.07 \\
\bottomrule
    \end{tabular}}
    \caption{The detection performance of ensembling $k$ modes (\textbf{DenseNet121} trained on \textbf{ImageNet-1K}) on each OoD data set \textit{w.r.t} different types of OoD detectors. The results with the {\bf lowest variances} are highlighted with bold fonts.}
    \label{app-tab:exp-ensemble-imgnet-dn121}
\end{table*}

\end{document}